\documentclass{ericsson}

\usepackage[english]{babel}

\usepackage{graphicx}
\usepackage{algorithmic}
\usepackage{algorithm}
\usepackage{svg}
\usepackage{float}
\usepackage{siunitx}
\usepackage{tabularray}
\UseTblrLibrary{booktabs}

\usepackage[nolist, printonlyused, nohyperlinks]{acronym}

\usepackage{subcaption}

\title{Generalization in Reinforcement Learning for Radio Access Networks}
\author[1]{Burak Demirel}
\author[1]{Yu Wang}
\author[1]{Cristian Tatino}
\author[1]{Pablo Soldati}
\affiliation[1]{Ericsson AB, Kista, Sweden}
\date{\today}
\correspondence{\email{burak.demirel@ericsson.com}}

\begin{document}

\begin{abstract}
Modern \acp{RAN} operate in highly dynamic and heterogeneous environments, where hand-tuned, rule-based \ac{RRM} algorithms frequently underperform. While \ac{RL} can surpass these heuristics in constrained scenarios, the unpredictable nature of radio channels and the diversity of cell deployments introduce substantial generalization challenges. Data-driven policies often overfit to their training distributions, leading to degraded performance when applied to unseen conditions.
To address this, we propose a generalization-focused \ac{RL} framework for \ac{RAN} control that: (i) robustly reconstructs dynamically varying states from partial and noisy observations, while encoding static and semi-static information—such as radio nodes, cell attributes, and their topology—through graph representations; (ii) applies extensive domain randomization to broaden the training distribution; and (iii) distributes data generation across multiple actors while centralizing training in a cloud-compatible architecture aligned with O-RAN principles. Although training generalizable policies increases computational and data-management complexity, our distributed design mitigates these costs by scaling data collection and training across heterogeneous network conditions.
Applied to downlink \ac{LA} across five \ac{5G} benchmarks, the resulting generalized \ac{RL} policy achieves approximately 10\% higher average cell throughput and spectral efficiency than the state-of-the-art \ac{LA} baseline in full-buffer \ac{MIMO} and \ac{mMIMO} scenarios, and approximately 20\% under high-mobility conditions. Furthermore, it matches the performance of specialized \ac{RL} policies in full-buffer traffic while delivering up to 4× and 2× throughput gains in \ac{eMBB} and mixed-traffic benchmarks, respectively. In larger deployments with nine cells, \ac{GAT} models achieve an additional 30\% throughput gain over \ac{MLP} baselines, underscoring the effectiveness of attention-based architectures in larger network settings.
These performance gains, combined with the scalable and distributed learning architecture, provide a practical foundation for AI-native \ac{6G} \acp{RAN}, enabling the deployment of a single, generalizable \ac{RL} agent that operates network-wide and consistently outperforms traditional heuristics.
\end{abstract}

\maketitle


\acrodefplural{MDP}[MDPs]{Markov decision processes}
\acrodefplural{RTG}[RTGs]{returns-to-go}

\begin{acronym}
  \acro{AI}{artificial intelligence}
  \acro{ACK}{positive acknowledgment}
  \acro{ARQ}{automatic repeat request}
  \acro{BE}{baseline encoding}
  \acro{BC}{behavioral cloning}
  \acro{BCQ}{batch-constrained deep Q-learning}
  \acro{CCTR}{Channel-Conditioned Target Return}
  \acro{CNN}{convolutional neural network}
  \acro{CPU}{central processing unit}
  \acro{CQL}{conservative Q-learning}
  \acro{CT}{continuous time}
  \acro{CV}{computer vision}
  \acro{DAVG}{discounted average}
  \acro{DLLA}{downlink link adaptation}
  \acro{DME}{data management entity}
  \acro{DP}{dynamic programming}
  \acro{DQN}{deep Q-network}
  \acro{DT}{decision transformer}
  \acro{eMBB}{enhanced mobile broadband}
  \acro{FB}{full buffer}
  \acro{ILLA}{inner-loop link adaptation}
  \acro{gNB}{next generation NodeB}
  \acro{gNB-CU}{gNB centralized unit}
  \acro{gNB-DU}{gNB distributed unit}
  \acro{GAT}{graph attention network}
  \acro{GNN}{graph neural network}
  \acro{GCN}{graph convolutional network}
  \acro{GPI}{generalized policy iteration}
  \acro{GPU}{graphics processing unit}
  \acro{HPC}{high-performance computing}
  \acro{LA}{link adaptation}
  \acro{LN}{layer normalization}
  \acro{LCM}{life cycle management}
  \acro{LSF}{load~sharing~facility}
  \acro{LSTM}{long short-term memory}
  \acro{LT}{learnable time}
  \acro{MAB}{multi-armed bandits}
  \acro{MC}{Monte Carlo}
  \acro{MDP}{Markov decision process}
  \acro{ML}{machine learning}
  \acro{MLP}{multi-layer perceptron}
  \acro{mMIMO}{massive MIMO}
  \acro{mMTC}{massive machine-type communications}
  \acro{NACK}{negative acknowledgment}
  \acro{NF}{network function}
  \acro{NLP}{natural language processing}
  \acro{OLLA}{outer-loop link adaptation}
  \acro{ORAN}{Open RAN}
  \acro{PE}{positional encoding}
  \acro{PHY}{physical layer}
  \acro{RL}{reinforcement learning}
  \acro{RLC}{radio link control}
  \acro{RNN}{recurrent neural network}
  \acro{RTGs}{returns-to-go}
  \acro{RTG}{return-to-go}
  \acro{RT RIC}{real-time RAN intelligent controller}
  \acro{near-RT RIC}{near-real-time RAN intelligent controller}
  \acro{RvS}{reinforcement learning via supervised learning}
  \acro{SACo}{state-action coverage}
  \acro{SADCo}{state-action density coverage}
  \acro{SE}{spectral efficiency}
  \acro{SMO}{service management and orchestration}
  \acro{SMOF}{SMO function}
  \acro{TBS}{transport block size}
  \acro{TD}{temporal difference}
  \acro{TQ}{relative trajectory quality}
  \acro{URLLC}{ultra-reliable low-latency communications}
  \acro{ZSG}{zero-shot generalization}
  \acro{VAE}{variational auto-encoder}
  \acro{2G}{Second Generation}
  \acro{3G}{3$^\text{rd}$~Generation}
  \acro{3GPP}{3$^\text{rd}$~Generation Partnership Project}
  \acro{4G}{4$^\text{th}$~Generation}
  \acro{5G}{5$^\text{th}$~Generation}
  \acro{6G}{6$^\text{th}$~Generation}
  \acro{AA}{Antenna Array}
  \acro{AC}{Admission Control}
  \acro{AD}{Attack-Decay}
  \acro{ADSL}{Asymmetric Digital Subscriber Line}
	\acro{AHW}{Alternate Hop-and-Wait}
  \acro{AMC}{Adaptive Modulation and Coding}
  \acro{AoA}{angle of arrival}
	\acro{AP}{Access Point}
  \acro{APA}{Adaptive Power Allocation}
  \acro{AR}{autoregressive}
  \acro{ARMA}{Autoregressive Moving Average}
  \acro{ATES}{Adaptive Throughput-based Efficiency-Satisfaction Trade-Off}
  \acro{AWGN}{additive white Gaussian noise}
  \acro{BB}{Branch and Bound}
  \acro{BD}{Block Diagonalization}
  \acro{BER}{bit error rate}
  \acro{BF}{Best Fit}
  \acro{BLER}{block error rate}
  \acro{BPC}{Binary power control}
  \acro{BPSK}{Binary Phase-Shift Keying}
  \acro{BPA}{Best \ac{PDPR} Algorithm}
  \acro{BRA}{Balanced Random Allocation}
  \acro{BCRB}{Bayesian Cram\'{e}r-Rao Bound}
  \acro{BS}{base station}
  \acro{CAP}{Combinatorial Allocation Problem}
  \acro{CAPEX}{Capital Expenditure}
  \acro{CBF}{Coordinated Beamforming}
  \acro{CBR}{Constant Bit Rate}
  \acro{CBS}{Class Based Scheduling}
  \acro{CC}{Congestion Control}
  \acro{CDF}{cumulative distribution function}
  \acro{CDMA}{Code-Division Multiple Access}
  \acro{CL}{Closed Loop}
  \acro{CLPC}{Closed Loop Power Control}
  \acro{CNR}{Channel-to-Noise Ratio}
  \acro{CPA}{Cellular Protection Algorithm}
  \acro{CPICH}{Common Pilot Channel}
  \acro{CoMP}{Coordinated Multi-Point}
  \acro{CQI}{channel quality indicator}
  \acro{CRB}{Cram\'{e}r-Rao Bound}
  \acro{CRM}{Constrained Rate Maximization}
	\acro{CRN}{Cognitive Radio Network}
  \acro{CS}{Coordinated Scheduling}
  \acro{CSI}{channel state information}
  \acro{CSIR}{channel state information at the receiver}
  \acro{CSIT}{channel state information at the transmitter}
  \acro{CUE}{cellular user equipment}
  \acro{D2D}{device-to-device}
  \acro{DCA}{Dynamic Channel Allocation}
  \acro{DE}{Differential Evolution}
  \acro{DFT}{Discrete Fourier Transform}
  \acro{DIST}{Distance}
  \acro{DL}{downlink}
  \acro{DMA}{Double Moving Average}
	\acro{DMRS}{demodulation reference signal}
  \acro{D2DM}{D2D Mode}
  \acro{DMS}{D2D Mode Selection}
  \acro{DPC}{Dirty Paper Coding}
  \acro{DRA}{Dynamic Resource Assignment}
  \acro{DSA}{Dynamic Spectrum Access}
  \acro{DSM}{Delay-based Satisfaction Maximization}
  \acro{ECC}{Electronic Communications Committee}
  \acro{EFLC}{Error Feedback Based Load Control}
  \acro{EI}{Efficiency Indicator}
  \acro{eNB}{Evolved Node B}
  \acro{EPA}{Equal Power Allocation}
  \acro{EPC}{Evolved Packet Core}
  \acro{EPS}{Evolved Packet System}
  \acro{ESPRIT}{estimation of signal parameters via rotational invariance}
  \acro{E-UTRAN}{Evolved Universal Terrestrial Radio Access Network}
  \acro{ES}{Exhaustive Search}
  \acro{FDD}{frequency division duplexing}
  \acro{FDM}{Frequency Division Multiplexing}
  \acro{FER}{Frame Erasure Rate}
  \acro{FF}{Fast Fading}
  \acro{FIM}{Fisher information matrix}
  \acro{FSB}{Fixed Switched Beamforming}
  \acro{FST}{Fixed SNR Target}
  \acro{FTP}{File Transfer Protocol}
  \acro{GA}{Genetic Algorithm}
  \acro{GBR}{Guaranteed Bit Rate}
  \acro{GLR}{Gain to Leakage Ratio}
  \acro{GOS}{Generated Orthogonal Sequence}
  \acro{GPL}{GNU General Public License}
  \acro{GRP}{Grouping}
  \acro{HARQ}{hybrid automatic repeat request}
  \acro{HMS}{Harmonic Mode Selection}
  \acro{HOL}{Head Of Line}
  \acro{HSDPA}{High-Speed Downlink Packet Access}
  \acro{HSPA}{High Speed Packet Access}
  \acro{HTTP}{HyperText Transfer Protocol}
  \acro{ICMP}{Internet Control Message Protocol}
  \acro{ICI}{Intercell Interference}
  \acro{ID}{Identification}
  \acro{ISAC}{integrated sensing and communication}
  \acro{IEEE}{Institute of Electrical and Electronics Engineers}
  \acro{IETF}{Internet Engineering Task Force}
  \acro{ILP}{Integer Linear Program}
  \acro{JRAPAP}{Joint RB Assignment and Power Allocation Problem}
  \acro{UID}{Unique Identification}
  \acro{IID}{Independent and Identically Distributed}
  \acro{IIR}{Infinite Impulse Response}
  \acro{ILP}{Integer Linear Problem}
  \acro{IMT}{International Mobile Telecommunications}
  \acro{INV}{Inverted Norm-based Grouping}
  \acro{IoT}{Internet of Things}
  \acro{IP}{Internet Protocol}
  \acro{IPv6}{Internet Protocol Version 6}
  \acro{ISD}{Inter-Site Distance}
  \acro{ISI}{Inter Symbol Interference}
  \acro{ITU}{International Telecommunication Union}
  \acro{JOAS}{Joint Opportunistic Assignment and Scheduling}
  \acro{JOS}{Joint Opportunistic Scheduling}
  \acro{JP}{Joint Processing}
  \acro{JS}{Jump-Stay}
  \acro{KKT}{Karush-Kuhn-Tucker}
  \acro{L3}{Layer-3}
  \acro{LAC}{Link Admission Control}
  \acro{LC}{Load Control}
  \acro{LOS}{Line of Sight}
  \acro{LP}{Linear Programming}
  \acro{LS}{least squares}
  \acro{LTE}{Long Term Evolution}
  \acro{LTE-A}{LTE-Advanced}
  \acro{LTE-Advanced}{Long Term Evolution Advanced}
  \acro{M2M}{Machine-to-Machine}
  \acro{MAC}{Medium Access Control}
  \acro{MANET}{Mobile Ad hoc Network}
  \acro{MCS}{modulation and coding scheme}
  \acro{MDB}{Measured Delay Based}
  \acro{MDI}{Minimum D2D Interference}
  \acro{MF}{Matched Filter}
  \acro{MG}{Maximum Gain}
  \acro{MH}{Multi-Hop}
  \acro{MIMO}{multiple input multiple output}
  \acro{MINLP}{Mixed Integer Nonlinear Programming}
  \acro{MIP}{Mixed Integer Programming}
  \acro{MISO}{Multiple Input Single Output}
  \acro{MLE}{maximum likelihood estimator}
  \acro{MLWDF}{Modified Largest Weighted Delay First}
  \acro{MME}{Mobility Management Entity}
  \acro{MMSE}{minimum mean squared error}
  \acro{MOS}{Mean Opinion Score}
  \acro{MPF}{Multicarrier Proportional Fair}
  \acro{MRA}{Maximum Rate Allocation}
  \acro{MR}{Maximum Rate}
  \acro{MRC}{Maximum Ratio Combining}
  \acro{MRT}{Maximum Ratio Transmission}
  \acro{MRUS}{Maximum Rate with User Satisfaction}
  \acro{MS}{mobile station}
  \acro{MSE}{mean squared error}
  \acro{MSI}{Multi-Stream Interference}
  \acro{MTC}{Machine-Type Communication}
  \acro{MTSI}{Multimedia Telephony Services over IMS}
  \acro{MTSM}{Modified Throughput-based Satisfaction Maximization}
  \acro{MU-MIMO}{multiuser multiple input multiple output}
  \acro{MU}{multi-user}
  \acro{MUSIC}{multiple signal classification}
  \acro{NAS}{Non-Access Stratum}
  \acro{NB}{Node B}
  \acro{NE}{Nash equilibrium}
  \acro{NCL}{Neighbor Cell List}
  \acro{NLOS}{Non-Line of Sight}
  \acro{NMSE}{Normalized Mean Square Error}
  \acro{NORM}{Normalized Projection-based Grouping}
  \acro{NP}{Non-Polynomial Time}
  \acro{NR}{New Radio}
  \acro{NRT}{Non-Real Time}
  \acro{NSPS}{National Security and Public Safety Services}
  \acro{O2I}{Outdoor to Indoor}
  \acro{OFDMA}{orthogonal frequency division multiple access}
  \acro{OFDM}{orthogonal frequency division multiplexing}
  \acro{OFPC}{Open Loop with Fractional Path Loss Compensation}
	\acro{O2I}{Outdoor-to-Indoor}
  \acro{OL}{Open Loop}
  \acro{OLPC}{Open-Loop Power Control}
  \acro{OL-PC}{Open-Loop Power Control}
  \acro{OPEX}{Operational Expenditure}
  \acro{ORB}{Orthogonal Random Beamforming}
  \acro{JO-PF}{Joint Opportunistic Proportional Fair}
  \acro{OSI}{Open Systems Interconnection}
  \acro{PAIR}{D2D Pair Gain-based Grouping}
  \acro{PAPR}{Peak-to-Average Power Ratio}
  \acro{P2P}{Peer-to-Peer}
  \acro{PC}{Power Control}
  \acro{PCI}{Physical Cell ID}
  \acro{PDF}{Probability Density Function}
  \acro{PDPR}{pilot-to-data power ratio}
  \acro{PER}{Packet Error Rate}
  \acro{PF}{Proportional Fair}
  \acro{P-GW}{Packet Data Network Gateway}
  \acro{PL}{Pathloss}
  \acro{PPR}{pilot power ratio}
  \acro{PRB}{physical resource block}
  \acro{PROJ}{Projection-based Grouping}
  \acro{ProSe}{Proximity Services}
  \acro{PS}{Packet Scheduling}
  \acro{PSAM}{pilot symbol assisted modulation}
  \acro{PSO}{Particle Swarm Optimization}
  \acro{PZF}{Projected Zero-Forcing}
  \acro{QAM}{Quadrature Amplitude Modulation}
  \acro{QoS}{Quality of Service}
  \acro{QPSK}{Quadri-Phase Shift Keying}
  \acro{RAISES}{Reallocation-based Assignment for Improved Spectral Efficiency and Satisfaction}
  \acro{RAN}{radio access network}
  \acro{RAT}{Radio Access Technology}
  \acro{RATE}{Rate-based}
  \acro{RB}{resource block}
  \acro{RBG}{Resource block broup}
  \acro{REF}{Reference Grouping}
  \acro{RM}{Rate Maximization}
  \acro{RNC}{Radio Network Controller}
  \acro{RND}{Random Grouping}
  \acro{RRA}{Radio Resource Allocation}
  \acro{RRM}{radio resource management}
  \acro{RSCP}{Received Signal Code Power}
  \acro{RSRP}{Reference Signal Receive Power}
  \acro{RSRQ}{Reference Signal Receive Quality}
  \acro{RR}{Round Robin}
  \acro{RRC}{Radio Resource Control}
  \acro{RSSI}{Received Signal Strength Indicator}
  \acro{RT}{Real Time}
  \acro{RU}{Resource Unit}
  \acro{RUNE}{RUdimentary Network Emulator}
  \acro{RV}{Random Variable}
  \acro{SAC}{Session Admission Control}
  \acro{SCM}{Spatial Channel Model}
  \acro{SC-FDMA}{Single Carrier - Frequency Division Multiple Access}
  \acro{SD}{Soft Dropping}
  \acro{S-D}{Source-Destination}
  \acro{SDPC}{Soft Dropping Power Control}
  \acro{SDMA}{Space-Division Multiple Access}
  \acro{SER}{Symbol Error Rate}
  \acro{SES}{Simple Exponential Smoothing}
  \acro{S-GW}{Serving Gateway}
  \acro{SINR}{signal-to-interference-plus-noise ratio}
  \acro{SI}{Satisfaction Indicator}
  \acro{SIP}{Session Initiation Protocol}
  \acro{SISO}{single input single output}
  \acro{SIMO}{Single Input Multiple Output}
  \acro{SIR}{signal-to-interference ratio}
  \acro{SLNR}{Signal-to-Leakage-plus-Noise Ratio}
  \acro{SMA}{Simple Moving Average}
  \acro{SNR}{signal-to-noise ratio}
  \acro{SORA}{Satisfaction Oriented Resource Allocation}
  \acro{SORA-NRT}{Satisfaction-Oriented Resource Allocation for Non-Real Time Services}
  \acro{SORA-RT}{Satisfaction-Oriented Resource Allocation for Real Time Services}
  \acro{SPF}{Single-Carrier Proportional Fair}
  \acro{SRA}{Sequential Removal Algorithm}
  \acro{SRS}{Sounding Reference Signal}
  \acro{SSB}{synchronisation signal block}
  \acro{SU-MIMO}{single-user multiple input multiple output}
  \acro{SU}{Single-User}
  \acro{SVD}{Singular Value Decomposition}
  \acro{TCP}{Transmission Control Protocol}
  \acro{TDD}{time division duplexing}
  \acro{TDMA}{Time Division Multiple Access}
  \acro{TETRA}{Terrestrial Trunked Radio}
  \acro{TP}{Transmit Power}
  \acro{TPC}{Transmit Power Control}
  \acro{TTI}{transmission time interval}
  \acro{TTR}{Time-To-Rendezvous}
  \acro{TSM}{Throughput-based Satisfaction Maximization}
  \acro{TU}{Typical Urban}
  \acro{UE}{user equipment}
  \acro{UEPS}{Urgency and Efficiency-based Packet Scheduling}
  \acro{UL}{uplink}
  \acro{UMTS}{Universal Mobile Telecommunications System}
  \acro{URI}{Uniform Resource Identifier}
  \acro{URM}{Unconstrained Rate Maximization}
  \acro{UT}{user terminal}
  \acro{VR}{Virtual Resource}
  \acro{VoIP}{Voice over IP}
  \acro{WAN}{Wireless Access Network}
  \acro{WCDMA}{Wideband Code Division Multiple Access}
  \acro{WF}{Water-filling}
  \acro{WiMAX}{Worldwide Interoperability for Microwave Access}
  \acro{WINNER}{Wireless World Initiative New Radio}
  \acro{WLAN}{Wireless Local Area Network}
  \acro{WMPF}{Weighted Multicarrier Proportional Fair}
  \acro{WPF}{Weighted Proportional Fair}
  \acro{WSN}{Wireless Sensor Network}
  \acro{WWW}{World Wide Web}
  \acro{XIXO}{(Single or Multiple) Input (Single or Multiple) Output}
  \acro{ZF}{zero-forcing}
  \acro{ZMCSCG}{Zero Mean Circularly Symmetric Complex Gaussian}
\end{acronym}

\acresetall

\renewcommand{\contentsname}{Contents}
\tableofcontents
\addtocontents{toc}{\protect\setcounter{tocdepth}{3}}  


\section{Introduction}\label{sec:1_intro}

Wireless communication networks have rapidly evolved into increasingly complex and heterogeneous systems, underscoring the need for advanced automation and optimization. Managing modern \acp{RAN}, such as the \ac{5G} \ac{NR} system, requires solving large-scale, often combinatorial problems under strict latency constraints—particularly those encountered in \ac{RRM}. The highly dynamic and stochastic nature of the \ac{RAN} environment further challenges the effectiveness of classical rule-based heuristics, which struggle to adapt to rapidly changing network conditions and diverse service requirements. These limitations have prompted a shift toward adaptive, data-driven solutions.

\Ac{AI} has shown strong potential in addressing these challenges for the next generation of \acp{RAN} by replacing rule-based designs with \ac{AI}-native solutions for various \ac{RAN} functionalities~\cite{Calabrese:2018}, such as traffic prediction, mobility optimization, interference management, and more~\cite{Zhang:21, Basaran:22, Ashour:23}. Among \ac{AI} methods, \ac{RL} has emerged as a new paradigm for solving complex decision-making problems in modern \ac{RAN} systems~\cite{Calabrese:2018}. By learning optimal control policies through interaction with the environment, \ac{RL} agents can adapt to dynamic conditions and optimize long-term performance in high-dimensional and uncertain settings. When combined with the abundance of data in \ac{RAN} applications, \ac{RL} presents a unique opportunity to uncover intricate data patterns and derive control policies that outperform human-crafted heuristics. This makes it particularly attractive for \ac{RRM} tasks—such as power control, handover, and interference management (see, e.g.,~\cite{Ghadimi:17, LHG+:19, CCK+:21, LZZ+:17})—as well as for network orchestration and management. This versatility is central to the vision of an \ac{AI}-native \ac{6G} \ac{RAN}, promoted by standardization bodies such as \ac{3GPP} and \ac{ORAN}, where RL-based applications deployed as xApps or rApps at various levels of the \ac{6G} architecture could support both dynamic real-time \ac{RRM} functions and slower service management and orchestration operations.

However, several studies~\cite{FMB:18, ZVM+:18, ZBP:18, GaG:19, SJT+:19, ICL+:19} have highlighted the limited generalization ability of \ac{RL} agents to transfer learned knowledge to novel environments. This limitation primarily arises from training algorithms within fixed environments, which results in overfitting to specific transition dynamics and reward structures~\cite{FMB:18}. Consequently, such models perform well under in-distribution conditions but suffer significant degradation when exposed to unseen environments. The challenge of \ac{RL} generalization is exacerbated in partially observable or uncertain \acp{MDP}, where incomplete knowledge or environmental uncertainty render the available states insufficient for optimal decision-making~\cite{ICL+:19}.

To generalize effectively, \ac{RL} agents must learn transferable representations that are robust to variations in the environment, including uncertainty and feature variability~\cite{KZG+:23}. While classical regularization techniques from supervised learning—such as weight decay, dropout, and batch or layer normalization—can enhance the robustness of \ac{RL} policies~\cite{CKH+:19}, domain randomization has recently emerged as a particularly effective alternative~\cite{TFR+:17}. This approach systematically introduces variability into training environments—through perturbations in state parameters or environmental dynamics—encouraging agents to learn robust policies that remain effective under such variations.

This technique is particularly valuable for sim-to-real (sim2real) transfer, where it helps \ac{RL} policies maintain performance when deployed in real-world environments, as demonstrated in robotics~\cite{ZWQ+:20}. In complex and dynamic domains such as \acp{RAN}, where live system training is often impractical or prohibitively costly, domain randomization facilitates robust sim2real transfer. A complementary approach is sim-to-sim (sim2sim) skill transfer~\cite{GKP+:19}, which enables \ac{RL} agents to transfer policies across simulated environments with varying characteristics, thereby enhancing generalization and increasing the likelihood of successful real-world deployment without repeated retraining~\cite{WAB:21}.

Despite extensive research into the use of \ac{RL} for \ac{RAN} operations, the challenge of achieving strong generalization performance in wireless communication systems remains underexplored~\cite{SGD+:25}. The unique characteristics of \acp{RAN}—including cell-specific environments, non-stationary radio channels, fluctuating interference levels, and traffic patterns, and the coexistence of multiple radio technologies—make this challenge particularly significant. \Ac{RL} policies trained under specific network conditions (e.g., data from one or a few cells) often generalize poorly to unseen scenarios, necessitating multiple specialized models (i.e., policies) and frequent retraining. This lack of generalization undermines the scalability and practicality of deploying \ac{RL} in real-world networks.

To address this challenge, this paper investigates generalization in \ac{RL} for \ac{RAN} applications, with a focus on improving adaptability and robustness under heterogeneous network conditions. The main contributions are listed as:
\begin{itemize}
    \item We define \ac{RL} generalization in the context of \acp{RAN}, emphasizing the distinctive challenges compared to classical \ac{RL} benchmarks.

    \item We identify three enablers of generalization in \ac{RL} for \ac{RAN}: (i) accurate reconstruction of state representations that capture the behavior of \ac{RAN} functions; (ii) increased diversity in training environments; and (iii) scalable distributed learning methods applicable to both simulated and operational networks.    

    \item We introduce a distributed learning architecture designed to address the dual challenge of ensuring data diversity and managing the computational complexity of training generalizable models. We show how this architecture supports large-scale data collection across simulated environments (for sim2sim transfer) and how its principles can extend to hierarchical, decentralized real-world \acp{RAN} to learn policies from field data.

    \item We develop a novel \ac{RL} algorithm for link adaptation that demonstrates the impact of the proposed enablers on improving generalization.

    \item We conduct extensive evaluations using a high-fidelity, system-level simulator compliant with \ac{5G} systems. Generalization is assessed via sim2sim transfer across environments with varying characteristics. Results show that the learned policies generalize to previously unseen radio conditions and outperform models fine-tuned to specific scenarios.
\end{itemize}

The paper is organized as follows. \Cref{sec:gen_in_rl_for_ran,sec:3_enablers} discuss the challenges and key enablers of \ac{RL} generalization in \ac{RAN} applications. \Cref{sec4:graphs} introduces graph models to enrich state representations with node attributes and topological information, while \Cref{sec:architecture,sec:integration2RAN} describe a distributed learning architecture designed to enhance \ac{RL} generalization in simulated and real-world \acp{RAN}, respectively. \Cref{sec:case_study} details the link adaptation case study, and \Cref{sec:RL_design} outlines the corresponding \ac{RL} design. Finally, \Cref{sec:numerical_examples} presents numerical evaluations, and \Cref{sec:conclusions} concludes the paper.


\section{RL Generalization in RAN Applications}\label{sec:gen_in_rl_for_ran}


\subsection{Definitions and Motivations}

Following~\cite{DeepLearning:15}, we define \ac{RL} generalization in the \ac{RAN} context as the ability of a policy to perform well under previously unseen network conditions, deployments, and radio environments. We focus in particular on \textit{\ac{ZSG}}~\cite{KZG+:23}, which aims to produce \ac{RL} policies that perform well in unseen \ac{RAN} environments \textit{without additional training}. This capability is critical because retraining for every environmental change (e.g., traffic pattern or network load variation) or deployment difference (e.g., different cells) is costly and impractical.

Complementary techniques such as \textit{few-shot adaptation} and \textit{transfer learning} can further facilitate \ac{RL} deployment in \ac{RAN} by enabling rapid adaptation from limited data or by leveraging knowledge from related scenarios. These methods are especially valuable in live systems, where \ac{RAN} vendors—who develop and train the models—may lack direct access to field data or large-scale deployments, limiting their ability to evaluate or retrain policies under realistic conditions.

Achieving \ac{ZSG} is also essential for scalable deployment of \ac{AI} models in \ac{RAN} systems. It enables network-wide replacement of a given RAN function with a single, generalized policy, ensuring consistent performance and reducing the need for continual retraining. In contrast, relying on cell-specific fine-tuned models would degrade robustness and severely hinder scalability due to the increased complexity of model lifecycle and data management. For example, maintaining separate models for thousands of radio cells for each \ac{AI}-driven \ac{RAN} function would be operationally burdensome and risk inconsistent behavior across the network.


\subsection{Challenges}

Key challenges in designing \ac{RL} agents that generalize well in \acp{RAN} include:

\begin{enumerate}
    \item \textbf{Heterogeneity of \ac{RAN} environments:} RAN deployments vary significantly in topology, scale, and radio access technologies. Differences in cell types (e.g., macro vs. small cells), hardware and deployment configurations (e.g., site location, antenna type and geometry), radio technologies (e.g., \ac{MIMO}, \ac{mMIMO}), as well as diversity of user devices and their distributions contribute to a highly diverse operational landscape. This heterogeneity poses major obstacles to generalization.

    \item \textbf{Stochasticity and partial observability:} Wireless channels are highly stochastic due to fading, interference, and random user activity. \ac{RL} agents often operate under partial observability, relying on local measurements without full network visibility. This complicates credit assignment and reduces the reproducibility of experience—particularly in multi-cell settings, where actions may have delayed or indirect effects. Learning robust policies under uncertainty remains a significant challenge.

    \item \textbf{Non-stationary network dynamics:} Wireless environments are inherently non-stationary. Factors such as user mobility, varying traffic loads, and time- and frequency-dependent channel conditions continually alter the environment’s transition and reward dynamics. Consequently, policies trained under specific traffic or mobility scenarios may degrade when deployed under sudden congestion, sparse demand, or changing user behavior. This temporal variability challenges static \ac{RL} models and requires adaptive mechanisms.
\end{enumerate}

Therefore, the \ac{RAN} environment poses significant challenges to \ac{RL} generalization, including highly variable operating conditions across time, space, and system configurations; partial and noisy observations; and frequent mismatches between training and deployment scenarios.


\section{Enablers of RL Generalization in RAN}\label{sec:3_enablers}

To promote \ac{RL} generalization for \ac{RAN} control functions, we identify three function-agnostic enablers: (i) robust state reconstruction (to handle partial and noisy observations), (ii) training-environment diversity (to cover a broad range of deployment scenarios), and (iii) distributed learning (to scale up data generation across many environments). In subsequent sections, we demonstrate their application to link adaptation.


\subsection{Robust State Reconstruction}

A prerequisite to attain model generalization over the \ac{RAN} environment is robustness to uncertainty, that is, the model's ability to cope with epistemic uncertainty arising from incomplete knowledge or limited data regarding the training domain~\cite{ICL+:19}. Epistemic uncertainty, for example, occurs when certain areas of the input space are poorly represented or completely missing in the training data, preventing the model from making confident predictions. To ensure robustness under uncertainty for \ac{RL} applications in \ac{RAN}, the input state of a model must be carefully constructed to accurately reflect environmental changes, even when the available observations are limited, delayed, or noisy.

\ac{RAN} functionalities typically operate under partial observability and heterogeneous information aging. Access nodes, for example, can only rely on local or incomplete views of the network state—either from direct sensing or from measurements reported by users or neighboring nodes. Furthermore, such information is inherently affected by uncertainties in its availability when a model needs it. For instance, while a model design may require certain measurements, such data may be temporarily unavailable, outdated, or unsupported, depending on the capability of specific network nodes or user devices. Moreover, different observations needed to reconstruct the state of a \ac{RAN} functionality may be gathered with very diverse timescales. As a result, the state reconstructed at a given time often contains features with different aging rates, leading to incomplete representations of the true state of the underlying system.

To enrich state reconstruction, we categorize information into \textit{static} (e.g., deployment topology, configuration parameters), \textit{semi-static} (or slowly varying, e.g., traffic load), and \textit{dynamic} (or rapidly changing, e.g., sub-second radio channel conditions). Dynamic information pertains to measurable state dynamics specific to the RAN functionality that \ac{RL} replaces. For example, for the link adaptation study case considered later, dynamic information includes path loss, channel state, and \ac{HARQ} feedback, and more. In contrast, static and semi-static information play a critical role in characterizing the radio environment, capturing the diversity of \ac{RAN} deployments and enabling models to operate effectively across the network. Such information may describe network topology, site locations, orientations, inter-site relationships, transmission technologies (e.g., antenna array type, carrier frequency, bandwidth, transmit power), and user device characteristics (e.g., chipset type, capabilities), providing the context needed for models to handle a wide variety of devices.

This classification allows us to consider different approaches to reconstruct the state of a \ac{RAN} function. The classic approach is to feed all information, indistinctly, to a \ac{DQN} model. However, to improve generalization across diverse RAN deployments and propagation environments, static and semi-static information can be processed using graph-based methods, such as \acp{GNN}~\cite{SGT:09}. Graph structures allow us to embed both radio access nodes and cell attributes, as well as their topological relationships, into rich state representations that can be transferred across different network deployments, improving model robustness to variations in network layout and environmental conditions~\cite{SAF:23, SZS:23, LLZ:25}. \Cref{sec4:graphs} further expands on this aspect.


\subsection{Training Environment Diversification}\label{sec3b:enabler_diversification}

An effective approach to achieving \ac{ZSG} in \ac{RL} is domain randomization~\cite{ZWQ+:20}, which deliberately varies aspects of the training environment so that agents learn robust, transferable policies. By exposing the agent to conditions often beyond those expected in deployment—through randomized user locations, traffic loads, channel conditions, and interference levels—the learned policy overfits less and generalizes better. In this way, domain randomization narrows the sim2real gap by making simulations sufficiently diverse to capture real-world uncertainty.

In the context of \acp{RAN}, domain randomization entails sampling heterogeneous network topologies, traffic and load scenarios, mobility patterns, and user densities. For example, an \ac{RL}–based scheduler could train on randomized combinations of traffic demands and cell loads to improve generalization to unpredictable deployment conditions. Prior work in robotics has demonstrated the benefits of zero-shot transfer~\cite{ZWQ+:20}, and we argue similar benefits for \ac{RL} in wireless networks, particularly when combined with domain adaptation techniques.

Despite its power, domain randomization poses three critical challenges:

\begin{enumerate}
    \item \textbf{Curse of dimensionality:} As more parameters (e.g., user positions, mobility models, fading profiles, antenna patterns) are randomized, the parameter space expands exponentially. Sampling this space adequately becomes computationally expensive or infeasible.

    \item \textbf{Unrealistic extremes:} Unconstrained randomization can yield non-physical scenarios—such as channel models with zero path loss or deployments involving hundreds of \acp{UE} within a \num{100}\nobreakdash-\si{\metre} cell—that misguide learning. To prevent this, domain knowledge and expert-defined constraints should be applied to keep randomization within physically plausible limits.

    \item \textbf{Training instability:} Domain randomization can introduce high variance in reward signals, destabilizing training and leading to non-convergent policies, excessive exploration, or catastrophic forgetting. This instability can be mitigated through curriculum learning~\cite{GBM+:17, MOC:20}, which gradually increases task complexity or environment randomness to ensure stable and progressive training.
\end{enumerate}


\subsection{Distributed Learning}

Distributed learning improves model robustness in sim2sim transfer by enabling parallel exploration of diverse environmental configurations. Multiple actors interact with distinct randomized environments concurrently, achieving broader domain coverage and reducing wall-clock time compared to sequential sampling. Architectures such as IMPALA~\cite{Espeholt:2018}, A3C~\cite{MBM+:16}, Ape-X~\cite{Horgan:2018}, and distributed PPO~\cite{openai2019dota2largescale} asynchronously update a shared policy from heterogeneous experiences, stabilizing training and mitigating overfitting to narrowly sampled conditions. A centralized replay buffer or gradient aggregator further integrates experiences across configurations, exposing the learner to a wider range of gradients and fostering robust, generalizable policies.

\Cref{sec:architecture} presents an architecture that scales distributed learning in \ac{RAN} simulators to leverage environmental diversity. \Cref{sec:integration2RAN} extends these design principles to real-world \ac{RAN} deployments, where data diversity stems from the collective experience of distributed network nodes performing policy evaluation concurrently during training.


\section{Graphs}\label{sec4:graphs}

\begin{figure*}[t]
    \centering
    \subfloat[\ac{UE}-centric network view.]{\includegraphics[width=0.5\columnwidth]{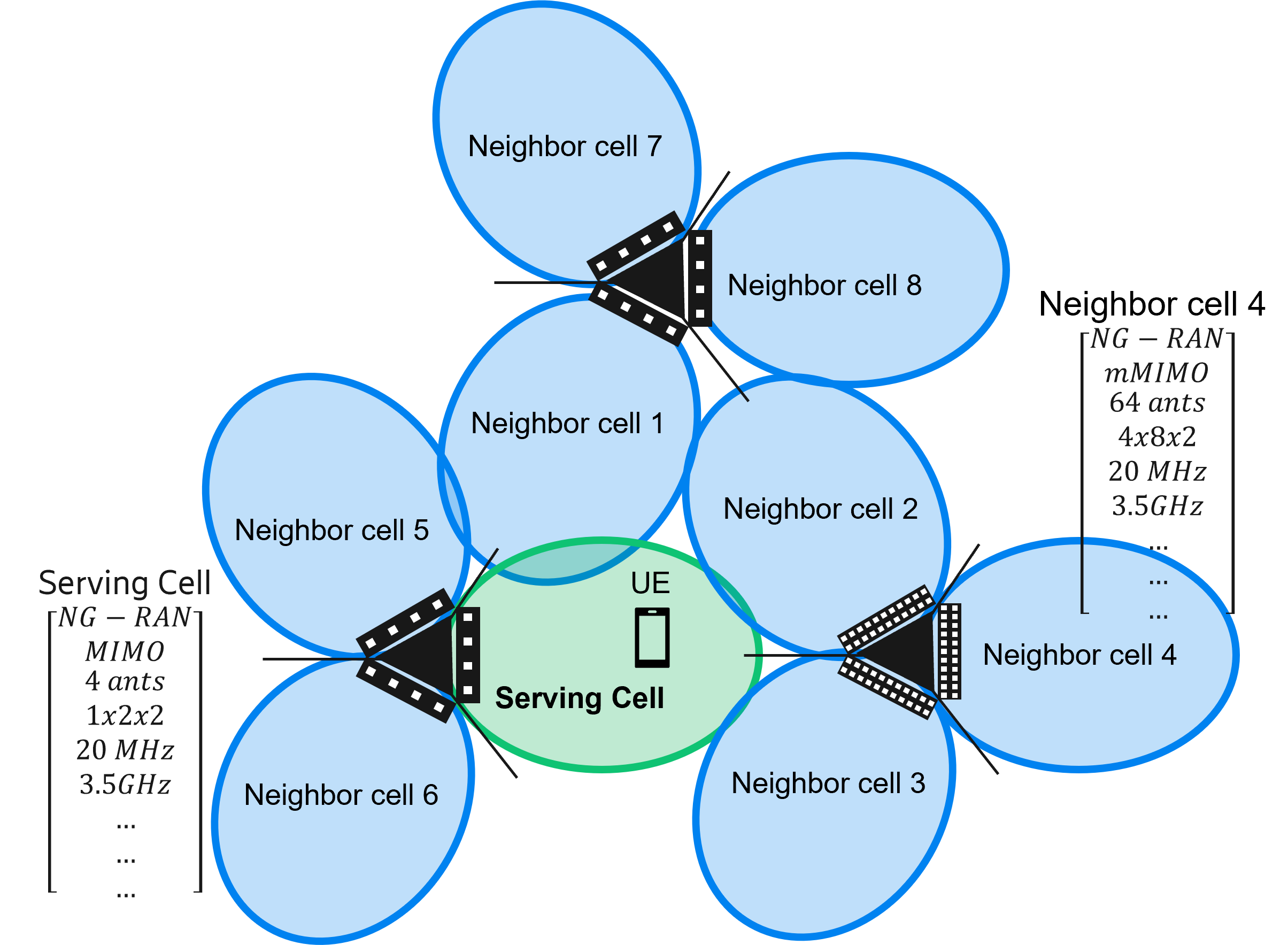}
    \label{fig:1a}}%
    \subfloat[\ac{UE}-centric graph.]{\includegraphics[width=0.5\columnwidth]{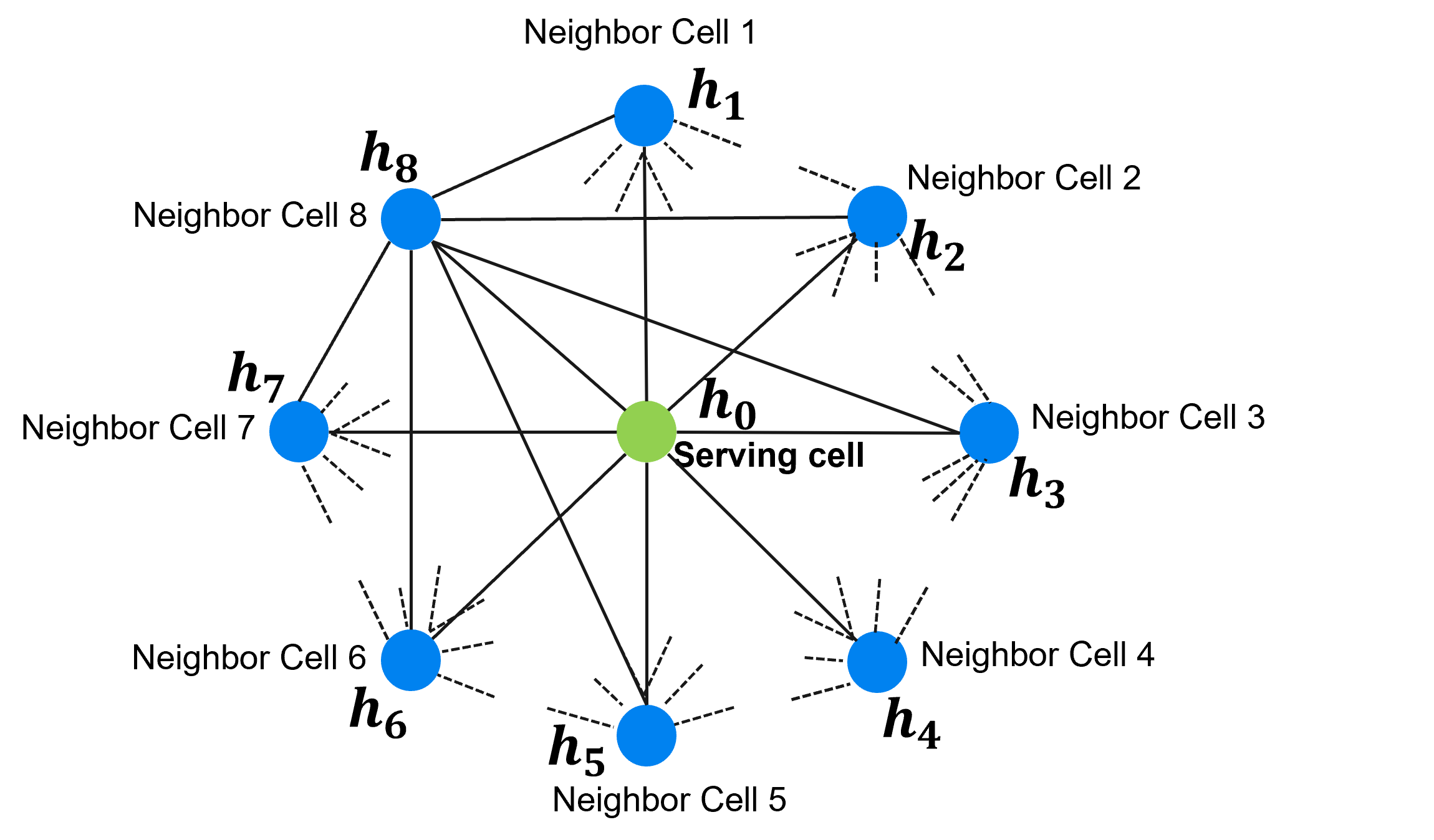}%
	\label{fig:1b}}
    \caption{\textbf{\ac{UE}-centric representation of multi-cell relationships in cellular networks.} (a) Radio environment snapshot as observed by a \ac{UE}, reporting reference signal measurements from its serving cell (green) and its eight strongest neighboring cells (blue), each characterized by antenna configuration, bandwidth, and frequency. (b) Abstracted graph model in which the \ac{UE}’s serving cell (central node, $h_{0}$) connects to neighboring cell nodes $h_{1}-h_{8}$. Edge thickness and node ordering represent measurement strength (e.g., RSRP), capturing inter-cell interference and handover affinities relevant to downstream machine learning tasks.
    }
    \label{fig:network_graph}
\end{figure*}

\Acp{GNN} extend traditional neural networks to graph-structured data, allowing the incorporation of connectivity information into the learning process. To achieve this, \acp{GNN} typically employ a feedforward architecture with multiple layers that iteratively aggregate and transform information characterizing the local neighborhood of a node using learnable permutation-invariant functions~\cite{SGT:09}. At each layer of a \ac{GNN}, a node updates its embedding by combining its own features with a summary of its neighbors’ features, typically through operations such as message passing and neighborhood pooling~\cite{SGT:09}. The number of layers determines how far information propagates in the graph: for instance, one layer captures immediate neighbors, two layers include two-hop neighbors, and so on. 

Several variants of \acp{GNN} have been proposed to capture different aspects of graph data, including:
\begin{itemize}
    \item \textbf{\Acp{GCN}}: Extend convolutional networks to graph structures, effectively capturing local node interactions~\cite{KW:17}.
    \item \textbf{\Acp{GAT}}: Incorporate attention mechanisms to weigh the importance of different neighbors during message passing~\cite{VCC:18}.
    \item \textbf{Graph transformers}: Leverage the self-attention mechanism of transformers to model long-range dependencies in graph-structured data, potentially capturing complex hierarchical relationships in \acp{RAN} systems~\cite{YCL:21}.
\end{itemize}

Graph representations, such as those learned by \acp{GNN}, are particularly well suited for modeling \ac{RAN} deployments, as they can capture both fine-grained local interactions (e.g., interference between nearby cells) and high-level global patterns (e.g., backbone connectivity or routing hierarchies)~\cite{LLZ:25}. 

We adopt this approach to learn expressive network embeddings from static and semi-static network information that captures both site/cell attributes and inter-site/cell relationships within a \ac{RAN} topology. Since \ac{GNN}-based models share aggregation functions across all nodes and edges, the same model can be reused—without retraining—for graphs with varying topologies that represent spatial and relational patterns of different portions of a \ac{RAN} deployment. Therefore, graph models can facilitate better \ac{RL} generalization in \ac{RAN} by improving the robustness and transferability of the model across diverse deployment scenarios. Furthermore, because the information characterizing a \ac{RAN} deployment is common across different functionalities, a learned embedding of the deployment can be reused in a variety of applications, including interference management, mobility management, and traffic prediction~\cite{SZS:23}.

Different graph representations of a \ac{RAN} deployment can be constructed by modeling base stations, antennas, or user devices as nodes, while edges can capture communication links, interference relationships, and geographical proximity. The graph structure in GNN-based models is often tailored to the specific use case. For example, \Cref{fig:network_graph} illustrates a graph representing a \ac{UE}-centric view of a \ac{RAN} neighborhood, where the central node represents the serving cell of a \ac{UE}, and surrounding nodes represent relevant interfering cells. Each node $i$ is associated with a feature vector $\mathbf{h}_i$ that characterizes the corresponding radio cell, while edges define proximity relationships between the associated cells. 

The approach shown in \Cref{fig:network_graph} is useful for designing \ac{RAN} functionalities that operate on \ac{UE}-specific parameters, such as power control, transmission beamforming, or link adaptation, where the functional state varies per \ac{UE}. In this case, the graph construction ensures that the context information provided to the \ac{GNN} reflects the portion of the network topology most relevant to each \ac{UE}—e.g., capturing relationships between surrounding cells that affect interference patterns, signal strength, and traffic conditions. In such a scenario, the serving cell can derive the graph instance using \ac{UE}-side measurements (e.g., \ac{RSRP} values of neighboring cells), while additional context information can be exchanged via standardized interfaces, such as the Xn interface in \ac{5G} \ac{NR} networks.

\Cref{sec:numerical_examples} describes how we integrate this approach with a \ac{DQN} design for the study case presented in~\Cref{sec:case_study}.


\section{Distributed Learning} \label{sec:architecture}

Since the introduction of the seminal \ac{DQN} algorithm~\citep{Mnih:2013}, several algorithmic and architectural advancements—such as GORILA~\citep{Nair:2015}, A3C~\citep{MBM+:16}, IMPALA~\citep{Espeholt:2018}, Ape-X~\citep{Horgan:2018}, R2D2~\citep{Kapturowski:2018}, Seed RL~\citep{Espeholt:2020}, NGU~\citep{BSV+:20}, and Agent57~\citep{BPK+:20}—have progressively improved the scalability and effectiveness of distributed \ac{RL} algorithms. Notably, IMPALA demonstrated how off-policy learning, combined with distributed actor-learner architectures, can effectively mitigate latency between action generation and gradient updates. These methods employ asynchronous updates, with distributed actors concurrently generating and evaluating training data across multiple parallel environments. Such architectures accelerate convergence through more frequent policy updates, increase sample diversity by learning from multiple environments concurrently, and improve final performance, as demonstrated in~\cite{Horgan:2018, Kapturowski:2018}. For example, the R2D2 architecture achieved up to a 20-fold increase in performance and a 50-fold reduction in training time compared to traditional \ac{DQN} implementations.


\subsection{Scaling Training in Network Simulators}

\begin{figure*}[!htbp]
	\centering
	\includegraphics[width=0.95\textwidth]{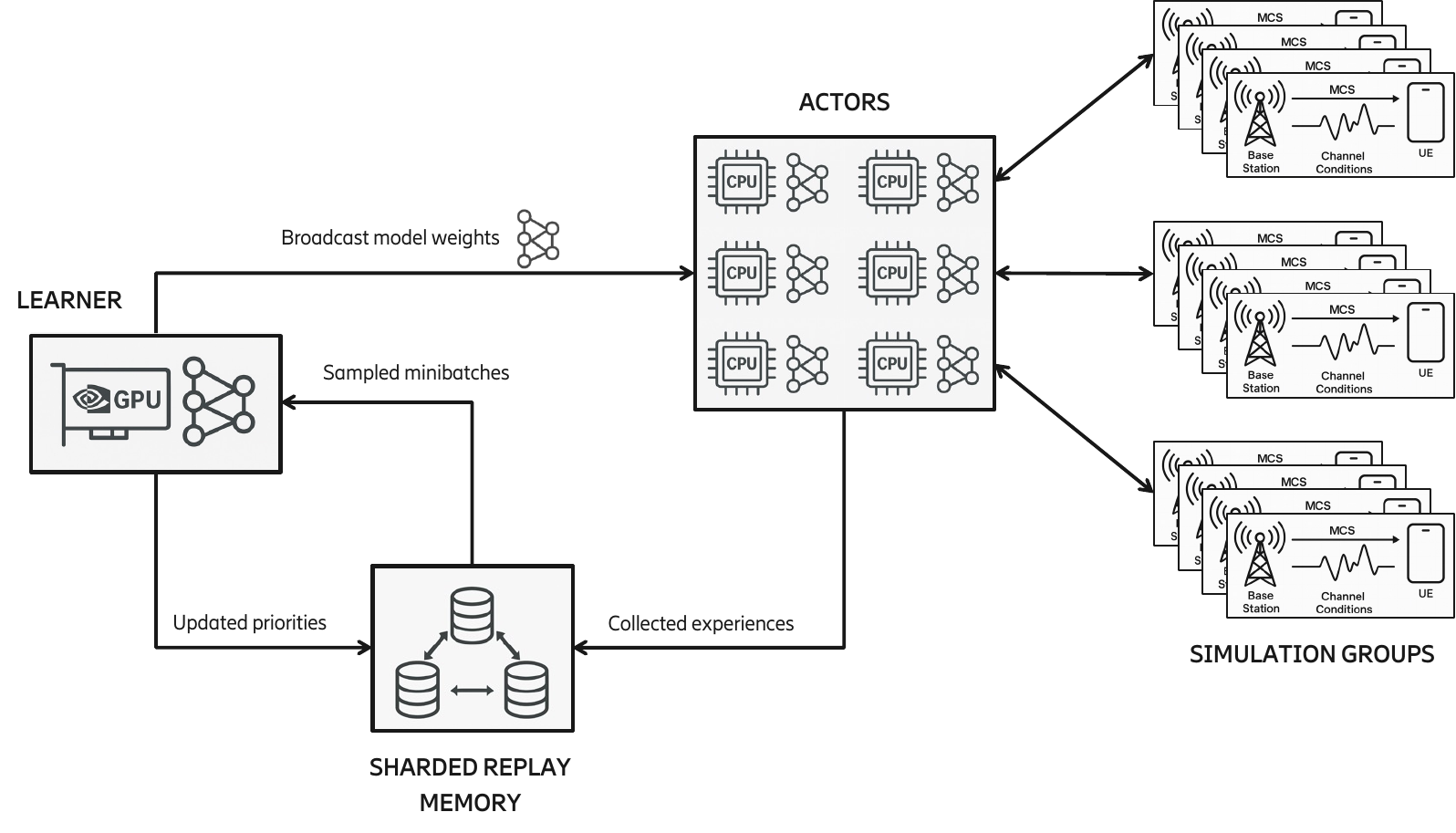}
    \caption{\textbf{Overview of a scalable architecture for a distributed \ac{RL} system integrated with wireless network simulators.} A GPU-based learner updates neural network weights using batches sampled from a distributed replay memory containing experience trajectories. Updated weights are periodically shared with a set of CPU-based actors, which concurrently interact with diverse simulation environments, emulating variable channel conditions and user behaviors. The resulting experience streams are pushed to the replay memory, with sampling priorities updated based on the learner feedback.}
	\label{fig:APEX}
\end{figure*}

In this work, we implement a large-scale distributed \ac{RL} algorithm, illustrated in~\Cref{fig:APEX}. The algorithm is based on the distributed prioritized experience replay architecture proposed in~\cite{Horgan:2018}, which decouples inference from learning, enables parallel inference, and significantly improves data collection capabilities. The architecture comprises three key components: a single learner operating on a \ac{GPU}; multiple independent actors running concurrently on \acp{CPU}; and a centralized, yet sharded, prioritized replay memory. A sharded replay memory partitions the experience replay buffer into several smaller segments, or shards, each storing and managing a portion of the collected experiences independently—often distributed across multiple threads, processes, or machines—to enhance scalability and throughput. Each actor concurrently interacts with a group of simulations hosted by different \acp{CPU}, further accelerating the data collection process. Furthermore, each actor employs a distinct exploration policy to enrich the diversity of data. Experiences generated by each actor through interactions with its environment instances are asynchronously contributed to the replay memory shard. 

The architecture is deployed on an \ac{HPC} cluster, where the learner, actors, and replay memory are co-located on a single compute node, while groups of simulation environments execute in parallel across multiple additional nodes. This setup enables high-throughput distributed \ac{RL} training with large-scale domain randomization over heterogeneous network simulations.

\subsection{Working Principles}

The interaction among actors, replay memory shards, and the learner—illustrated in~\Cref{fig:APEX}—proceeds as follows:

\textbf{Experience generation by actors:} Each actor runs a replica of the current policy network on its assigned \ac{CPU} core and interacts with multiple simulated environments, which are distributed across \ac{CPU} cores on different compute nodes. At each step, or after completing an episode rollout, the actor generates experience tuples of the form (state, action, reward, next state). These experiences are first stored in a local buffer and then sent in batches to the replay memory, along with initial priority values.

\textbf{Experience distribution to shards:} Incoming experiences are assigned to replay shards using one of the following strategies:

\begin{itemize}
    \item \textit{Round-robin or load balancing:} Experiences are distributed in a round-robin fashion or via a load-balancing mechanism to evenly fill all shards. For example, with four shards, an actor may send its first experience to shard 1, the next to shard 2, and so on cyclically. This strategy prevents bottlenecks and promotes uniform growth across shards.

    \item \textit{Fixed actor-to-shard mapping:} Alternatively, each actor—or group of actors—can be statically mapped to a specific shard, sending all experiences exclusively to that shard throughout training.
\end{itemize}

\textbf{Data storage in shards:} Each shard operates as an independent, typically circular, replay buffer. When full, a shard evicts older experiences or, in prioritized replay, replaces those with the lowest priority. Since shards are decoupled, data insertions to one shard do not interfere with others, enabling parallel writes.

\textbf{Sampling from shards by the learner:} The probability of sampling a transition \( e_{i,j} \) from the \( j^\text{th} \) shard \( \mathcal{R}_j \) is:

\begin{align}
\Pr(e_{i,j}) = \frac{p_{i,j}^\alpha}{\sum_j \sum_k p_{k,j}^\alpha} 
= \left( \frac{p_{i,j}^\alpha}{\sum_k p_{k,j}^\alpha} \right) 
\cdot \left( \frac{\sum_k p_{k,j}^\alpha}{\sum_j \sum_k p_{k,j}^\alpha} \right) 
= \Pr(e_{i,j} \mid \mathcal{R}_j) \cdot \Pr(\mathcal{R}_j). \label{eqn:sample_from_shard}    
\end{align}

The learner periodically queries each shard to retrieve its total priority and compute \( \Pr(\mathcal{R}_j) \), as defined in~\Cref{eqn:sample_from_shard}. It then samples a proportional number of experiences from each shard based on \( \Pr(e_{i,j} \mid \mathcal{R}_j) \), combines them, and shuffles to form a training batch. This strategy ensures batch diversity by drawing across all shards.

\textbf{Learning update:} The learner performs a gradient update on the neural network using the sampled batch. After each update, the learner independently updates the priorities of the sampled experiences.

\textbf{Priority updates:} Each sampled experience is associated with a priority, typically derived from the \ac{TD} error. After updating priorities, the learner sends these values to the corresponding shard. Each shard independently maintains its own data structure (e.g., segment tree or heap) to manage priorities and facilitate efficient sampling.

\textbf{Iterative process:} Actors continuously generate new experiences and send them to their designated shards, while the learner repeatedly samples from shards and updates the policy. This pipeline runs in parallel. To maintain policy consistency, the learner periodically broadcasts updated network weights to all actors. As a result, the replay memory contains a mixture of experiences collected under different policies, making the learning process inherently off-policy.


\section{Integrating Distributed Learning in RAN Systems} \label{sec:integration2RAN}

\begin{figure*}[t!]
    \centering
    \includegraphics[width=0.9\textwidth]{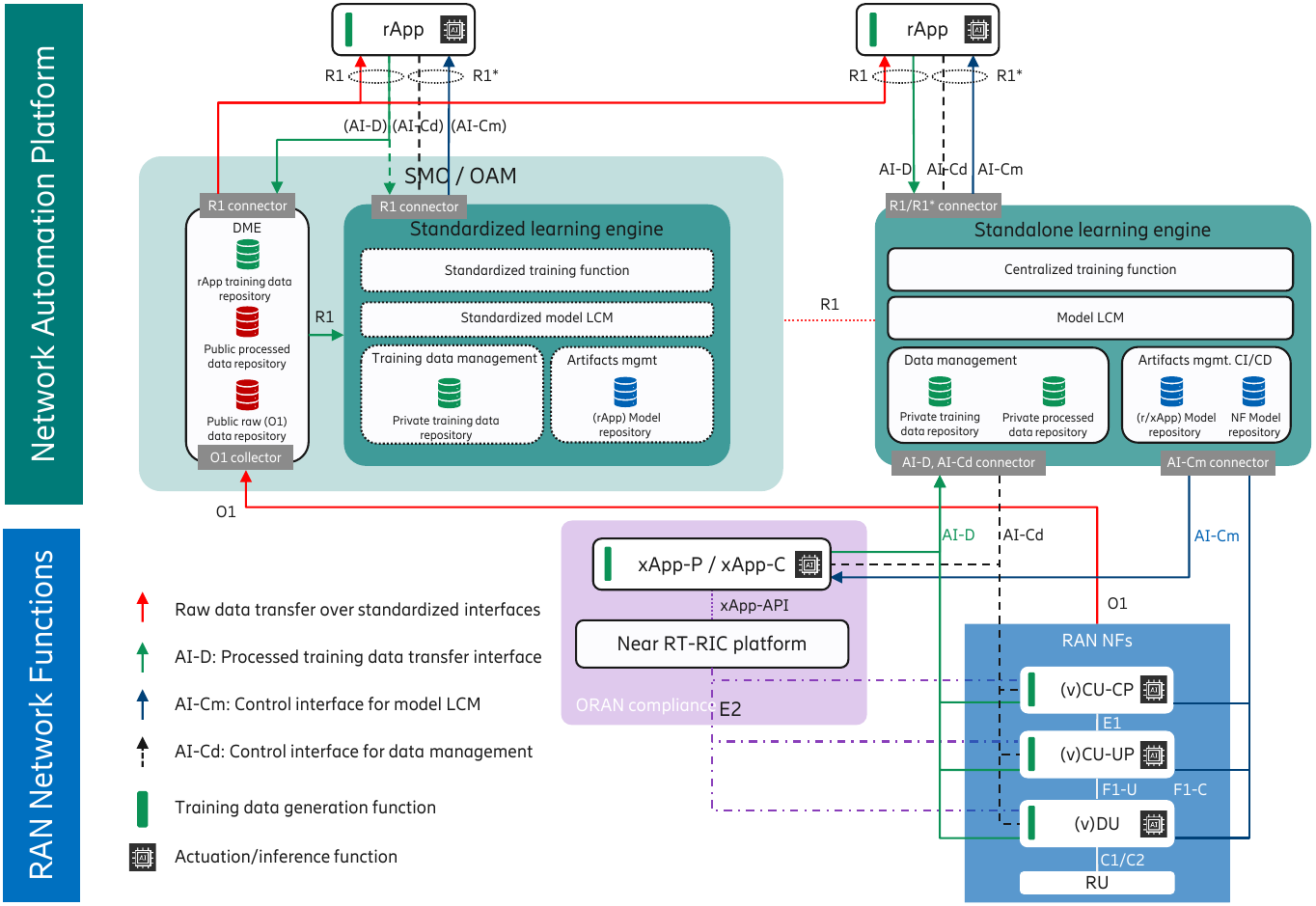}
    \caption{
        \textbf{An \ac{AI} architecture integrating distributed learning in a \ac{5G} \ac{RAN} system, with the learning engine realized either as a standardized function within the network automation layer or as a standalone support function.} A centralized learning engine performs model training, data management, and model lifecycle management, while distributed RAN network functions and r/xApps host inference (actor) functions that generate training data from local environments. Standardized and proprietary interfaces support the exchange of raw and processed data, control information, and model artifacts between the learning engine, SMO/OAM components, and RAN functions, enabling deployment across different layers of the RAN architecture.
        }
    \label{fig:AI_in_RAN_architecture}
\end{figure*}

The foundational design principles of \ac{RL} architectures—distributing policy evaluation (i.e., data generation) across actors while centralizing training and data management—align naturally with the hierarchical structure of \ac{RAN} systems. These principles support an \ac{AI} architecture in which a centralized learning engine orchestrates and provides support services to \ac{AI} applications deployed in distributed \ac{RAN} \acp{NF}, such as \acp{gNB-CU}, \acp{gNB-DU}, for various control and prediction tasks, or alternatively deployed as x/rApps in support of \ac{RAN} \acp{NF}, as illustrated in~\Cref{fig:AI_in_RAN_architecture}.

The architecture in~\Cref{fig:AI_in_RAN_architecture} enables distributed learning by allowing \ac{RAN} \acp{NF} that host \ac{AI}-driven functionalities to collect experiences from local environments and contribute to a shared dataset for training a generalized model. Analogous to domain randomization in simulations, distributing policy evaluation across \ac{RAN} \acp{NF} during training increases sample diversity, mitigates data bias, and enhances robustness against uncertainties inherent to real-world \acp{RAN}. Most notably, training models directly on field data can substantially mitigate performance discrepancies induced by the sim2real gap.

This architecture features a learning engine deployment that is agnostic to specific \ac{AI} applications, while the deployment of actors (or inference functions more broadly) is use-case dependent. For example, actors replacing physical layer (L1) and medium access control layer (L2) functions—such as resource scheduling or link adaptation—could be hosted in \acp{gNB-DU}. In contrast, actors replacing networking layer (L3) functions—such as load balancing or mobility handover—could reside in \acp{gNB-CU}. To support AI-native network orchestration functions—e.g., optimizing the coverage-capacity trade-off across a network area—actors may be realized as rApps. Alternatively, actors (or inference functions) may be deployed as xApps in the \ac{near-RT RIC} of an \ac{ORAN} system, providing insights or configuration updates to \ac{RAN} \acp{NF}.

To fulfill its central role and remain use-case agnostic, the learning engine can be deployed at higher layers of the \ac{RAN} architecture. A cloud-based implementation provides abundant and flexible compute and storage resources to deliver \ac{AI} services to distributed \ac{RAN} \acp{NF} underneath. The learning engine supports a variety of services, including model training, data services (e.g., data generation control, storage, and management), and model lifecycle management (LCM) operations that ensure model compliance with reliability, performance, security, ethical, and standards requirements. While LCM is centrally orchestrated, certain computational tasks—such as performance monitoring or drift detection—can be delegated to the \ac{RAN} \acp{NF} that execute \ac{AI} models locally. This approach concentrates computationally intensive tasks, such as training and large-scale data storage, where they are most needed, thereby promoting a cost-efficient and scalable integration of \ac{AI} into \ac{RAN} systems.

\Cref{fig:AI_in_RAN_architecture} presents two approaches to integrate a learning engine into a \ac{RAN} system: (a) as a standalone support function and (b) as a standardized support function. Although~\Cref{fig:AI_in_RAN_architecture} depicts both approaches for brevity, a practical deployment would typically adopt only one.

\Cref{fig:AI_in_RAN_architecture} exemplifies a standalone learning engine coexisting with an \ac{SMO} platform in the network automation layer, communicating with actors (inference functions) hosted in \ac{RAN} \acp{NF} via proprietary or extensions of standardized interfaces. The AI-D interface (shown in green) supports data collection from distributed \ac{RAN} \acp{NF}; AI-Cd (shown in black) orchestrates training data generation for specific \ac{AI} applications; and AI-Cm (shown in blue) supports artifact management and model deployment. A standalone approach offers maximum design flexibility, allowing \ac{RAN} \acp{NF} and r/xApps to locally process raw data into customized training samples that meet algorithmic requirements—such as state-action-reward tuples for \ac{RL}—prior to transferring them to the learning engine.

A standardized approach, on the other hand, relies on standardized interfaces and functional frameworks. Using the \ac{ORAN} terminology,~\Cref{fig:AI_in_RAN_architecture} illustrates a learning engine implemented as an \ac{SMOF}, relying on other \acp{SMOF}, such as a \ac{DME}, for data collection, management, and storage. This approach mandates the use of standardized protocols and formats, which may not always align with specific \ac{AI} design requirements.

For instance, \ac{RAN} \acp{NF} may transmit raw observations to a \ac{DME} over O1/O2 interfaces (shown in red) in accordance with interface specifications. However, raw observations may not be directly usable for training. The \ac{DME} must then process the data into training samples suited to the \ac{AI} application, which are subsequently transmitted—e.g., via the R1 interface (shown in green)—to the learning engine. The primary motivation for a standardized approach is to enable inter-vendor interoperability, allowing network operators to build architectures composed of components from different vendors. However, standardizing the learning engine’s functionalities may pose significant design constraints.

\section{Case Study: Link Adaptation}\label{sec:case_study}

\Ac{LA} is a core function of modern wireless communication systems that employs adaptive modulation and coding to maximize spectral efficiency. Specifically,~\ac{LA} optimizes the modulation order and code rate of a packet transmission to match the link capacity based on the underlying radio conditions, using receiver-side \ac{CSI} such as \ac{CQI}~\cite{3GPP38214}. The \ac{LA} parameters are then encoded into a unique value, known as the \ac{MCS} index, and sent to the receiver to facilitate packet decoding~\cite{3GPP38214}. 


\subsection{Outer-Loop Link Adaptation}

State-of-the-art communication systems, such as \ac{5G} \ac{NR}, rely on an \ac{OLLA} scheme~\cite{PMK+:07} to adjust \ac{SINR} estimates. These estimates are often inferred from outdated or imprecise \acp{CQI}—to maximize spectral efficiency while meeting a predefined \ac{BLER} target. \Ac{OLLA} typically operates as an integrator, adjusting the \ac{SINR} estimate based on \ac{HARQ} feedback—raising it after successful transmissions and lowering it following failures. The \ac{BLER} target defines the ratio between the increment and decrement steps~\cite{DTR+:15}, and the adjusted \ac{SINR} is subsequently used to select the most appropriate \ac{MCS} parameters. 

Although several variants of this approach have been proposed~\cite{BAT+:14, DTR+:15, PDH:15, CGT+:16, OhS:16, DLM+:17, NAO:21, RGL+:22, NID:23}, factors such as user mobility, traffic burstiness, channel aging, limited receiver-side information, and heterogeneous \ac{UE} populations (e.g., with varying receiver and hardware capabilities) make \ac{LA} a difficult control task to model~\cite{Chen:23}, rendering \ac{OLLA} suboptimal.


\subsection{Reinforcement Learning-Based Link Adaptation}


Recognizing the limitations of traditional heuristics, there is growing interest in developing alternative, proactive solutions to \ac{LA} by leveraging the predictive capabilities of \ac{ML}, and \ac{RL} in particular. Numerous studies have proposed \ac{LA} solutions for \ac{5G} \ac{NR} networks using various Q-learning variants. Notably,~\cite{LCV:12},~\cite{BMP:14}, and~\cite{MAN+:19} employed tabular Q-learning;~\cite{WTM:20},~\cite{Chen:23}, and~\cite{KHV+:22} utilized \ac{DQN}; while~\cite{GWH+:22} proposed an actor-critic RL algorithm for optimizing \ac{MCS} selection. Other work, such as~\cite{saxena:21} and~\cite{KCC:17}, has explored bandit-based approaches.

While these studies demonstrate that \ac{RL} can improve \ac{LA} performance over rule-based methods, they introduce new challenges that limit practical viability. For example, tabular Q-learning does not scale well with the state-action space of \ac{LA}, and bandit-based approaches often rely on overly simplified problem formulations that ignore retransmissions. Most notably, many studies train and evaluate \ac{RL} models using simplistic simulators, assuming ideal conditions and fixed environments, without considering whether the models can generalize effectively to the diverse conditions found across radio cells in a real-world, network-wide \ac{RAN} deployment.


\section{Modelling Link Adaptation as an MDP}\label{sec:RL_design} 

We jointly model the downlink \ac{LA} and \ac{HARQ} processes in \acp{RAN} as a \ac{MDP}, and present a novel \ac{DQN}-based design combined with the enablers described in~\Cref{sec:3_enablers} to demonstrate \ac{RL} generalization for \ac{RAN} applications.


\subsection{Episode Design}

Link adaptation and \ac{HARQ} operate on a per-\ac{UE} and per-packet transmission basis. We formulate this problem as an episodic \ac{MDP}~\citep{Sutton1998} defined by $\mathcal{M} = \left\langle\mathcal{S}, \mathcal{A}, p, r, \gamma, \rho_{0} \right\rangle$, where $\mathcal{S}$ denotes the state space, $\mathcal{A}$ the action space, $p(s^{\prime} \mid s, a)$ the transition dynamics, $r:\mathcal{S} \times \mathcal{A} \rightarrow \mathbb{R}$ the reward function, $\gamma \in [0,1)$ the discount factor, and $\rho_{0}$ the initial state distribution.

\begin{figure*}[t]
\centering
\includegraphics[width=0.95\linewidth]{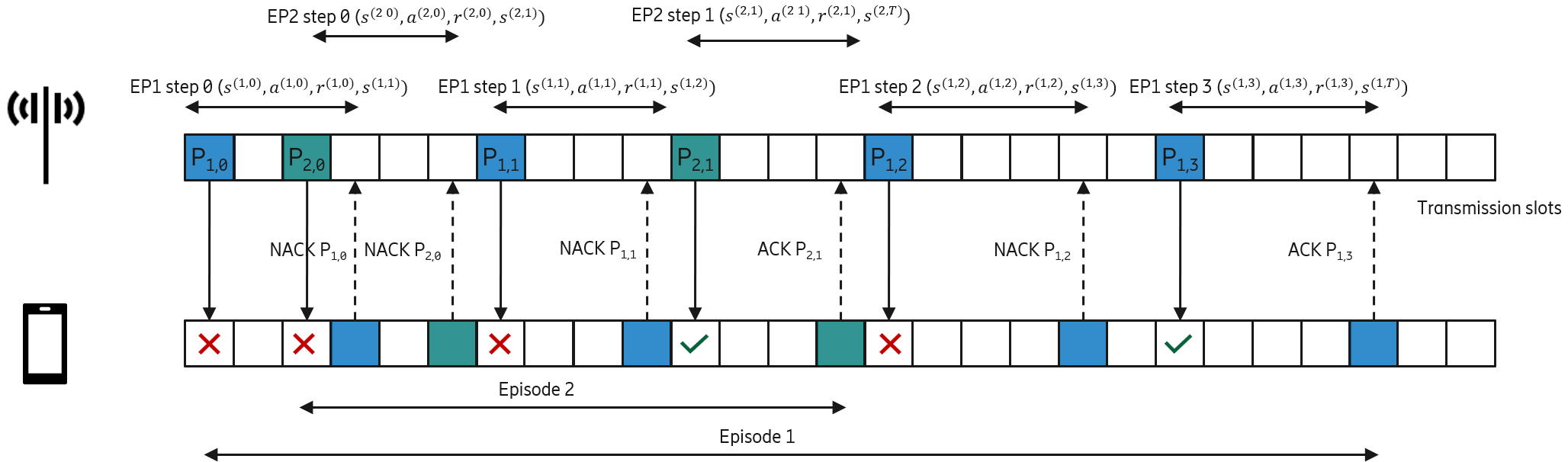}
\caption{\textbf{Markov decision process formulation for downlink link adaptation and HARQ in wireless communication.} The figure illustrates how episodes in an \ac{MDP} are structured to model the interaction between the \ac{BS} and \ac{UE} during downlink transmission under a hybrid \ac{HARQ} protocol. Each episode corresponds to the transmission of a single packet and comprises a sequence of steps, each defined by a state~($s_t$), an action~($a_t$), and a reward~($r_t$). The process begins with the arrival of a new packet at the \ac{BS} (shown in dark blue) and proceeds through repeated transmissions, each followed by either a \ac{NACK} or an \ac{ACK} from the \ac{UE}. Failed transmissions are marked in red, and successful transmissions in green. The corresponding \ac{NACK} and \ac{ACK} signals are indicated in light red and light green, respectively. Light blue segments represent retransmissions of the same packet due to decoding failures, repeated until successful reception is confirmed via an \ac{ACK}.}
\label{fig:episode}
\end{figure*}

An \ac{MDP} episode models the lifespan of a \ac{UE} packet in the \ac{HARQ} process—from its first transmission to either a successful reception or the packet being dropped upon $N$ transmission attempts, as illustrated in~\Cref{fig:episode}. This enables us to train a single \ac{RL} policy from the collective experience generated by any \acp{UE} across the network. Each step in the episode represents the duration of a packet transmission in the \ac{HARQ} process, from the selection of LA parameters (i.e., the \ac{RL} action) to the reception of the associated \ac{HARQ} feedback, i.e., a \ac{ACK} or a \ac{NACK} for successful or failed transmission, respectively. For instance, the \ac{3GPP} \ac{5G} \ac{NR} system, used in our evaluations, supports at most four packet retransmissions. Hence, the episode length $N$ may range from one to five steps. Each step is characterized by a state, an action, and an associated reward, as presented next. 

\subsection{Reward Design}

Since the goal of \ac{LA} is to adapt the \ac{MCS} parameters to channel conditions varying in time and frequency, thereby maintaining high link spectral efficiency, we propose a reward design that directly reflects the spectral efficiency of individual transmissions:
\begin{equation}\label{eqn:reward}
r_n = SE_n \cdot \delta_n - \alpha \cdot n \cdot (1-\delta_n), \quad n\in\{1, \dots, N\},
\end{equation}
with
\begin{equation}\label{eqn:ack_nack}
\delta_n =
\begin{cases}
1 & \text{if ACK at the $n$-th transmission attempt},\\
0 & \text{if NACK at the $n$-th transmission attempt},
\end{cases}
\end{equation}
where $n$ is a packet transmission attempt counter, $SE_n$ denotes the spectral efficiency achieved at attempt $n$, and $\alpha \in \mathbb{R}_{+}$ is a robustness weight penalizing retransmissions.

Optimizing exclusively for spectral efficiency may lead to excessive retransmissions, consequently lowering the effective throughput, which considers the total time required to deliver a packet. Rule-based \ac{LA} algorithms, such as \ac{OLLA} (see~\Cref{sec:case_study}), address this issue by enforcing adherence to a predefined \ac{BLER} target. Conversely, the reward formulation in~\Cref{eqn:reward} incorporates two complementary mechanisms that encourage policies not to exploit retransmissions excessively, balancing spectral efficiency against throughput. First, when a packet is successfully transmitted (\ac{ACK}) at the $n$-th attempt, $SE_n$ represents the spectral efficiency accounting for the cumulative time-frequency resources utilized across all $n$ transmission attempts. Second, each failed packet transmission (\ac{NACK}) incurs a penalty proportional to the transmission attempt number $n$, scaled by $\alpha$.

Furthermore, the reward design in~\Cref{eqn:reward} is agnostic to the underlying radio transmission technology, scheduling policy, and system bandwidth. This key property facilitates model generalization across heterogeneous \ac{RAN} environments, enabling the deployment of a single model across nodes operating at bandwidths ranging from a few to hundreds of MHz. Specifically, formulating the reward in terms of spectral efficiency makes it independent of the \ac{TBS} (i.e., the number of information bits in a packet) and the number of resource elements ($N_{RE}$) allocated by the scheduler. Unlike the wide-ranging values of \ac{TBS} and $N_{RE}$, spectral efficiency values are standardized and bandwidth-independent, as exemplified by the \ac{MCS} tables defined in the \ac{3GPP} NR specifications~\cite{3GPP38214}. Additionally, measuring spectral efficiency on a per-spatial-layer basis enables the training of a unified model applicable to both single- and multi-layer \ac{MIMO} transmissions.


\subsection{Action Space Design}

The action space of the \ac{LA} problem consists of a discrete set of \ac{MCS} indices, represented by $\mathcal{A}=\{0, \dots, M-1\}$, each corresponding to distinct data rates defined by the communication standard. For instance, in our evaluations, we utilize the \ac{MCS} indices specified by the \ac{3GPP} \ac{5G} \ac{NR} specifications (e.g., Tables 5.1.3.1-1 and 5.1.3.1-2 in~\cite{3GPP38214}, for modulation orders up to 64QAM and 256QAM, respectively).

Given a state $s \in \mathcal{S}$, an action $a_m = m$ implicitly determines the modulation order, code rate, and the targeted link spectral efficiency for a packet transmission. Together with the allocated time-frequency resource blocks assigned to the \ac{UE}, this action allows us to calculate the \ac{TBS} payload for the first transmission of a packet, e.g., based on \ac{NR} specifications~\cite{3GPP38214,3GPP38211}. The resulting \ac{TBS} is subsequently used to compute the corresponding reward, as described by~\Cref{eqn:reward}. For retransmissions, however, the \ac{TBS} remains unchanged from the initial packet transmission, while the modulation order and code rate may vary.


\subsection{State Space Design}\label{sec:7d}

To enable distributed learning of a single \ac{RL} policy from the collective experience of all \acp{UE} across the network, we model the state of the \ac{LA} process by combining (a) a \ac{UE}-centric view of the \ac{RAN} environment surrounding the \ac{UE}, consisting of semi-static information characterizing network nodes, radio cells, and the \ac{UE} itself (see~\Cref{sec4:graphs}), and (b) dynamic information characterizing relevant features of the \ac{LA} process. Together with domain randomization during training (see~\Cref{sec:3_enablers}), this allows us to train models that generalize across the \ac{RAN} environment.


\subsubsection{Semi-static Information} \label{subsec7:static_information}

We describe the network deployment surrounding a \ac{UE} by considering its serving cell and a subset of the most relevant interfering cells, as illustrated in~\Cref{fig:1a}. Specifically, we select $K_I$ cells whose signal strength, measured by the \ac{UE}, exceeds a predefined threshold. Thus, the composition of surrounding cells may change over time depending on \ac{UE} mobility. For each cell, we include semi-static configuration information such as site type (e.g., LTE or \ac{NR}), carrier frequency, antenna array configuration (e.g., \ac{MIMO} or \ac{mMIMO}), geometric properties (e.g., inter-site distance), bandwidth, and transmit power.

Since the strength and quality of signals perceived by a \ac{UE} can vary significantly due to manufacturing characteristics—such as chipset, antenna hardware, and algorithms for signal processing or channel state estimation—the \ac{MDP} state should capture information characterizing different \ac{UE} types to generalize effectively across the diversity of the \ac{UE} population. To this end, we include indicators representing the \ac{UE} type and antenna array size.

We feed this information to either a \ac{DQN} or a graph-based model, as described in~\Cref{sec4:graphs}. While a limited set of semi-static parameters is sufficient to model the \ac{RAN} deployment in our simulations (cf.~\Cref{sec:numerical_examples}), additional parameters may be beneficial to better characterize the diversity of real-world network deployments. 


\subsubsection{Dynamic Information} \label{subsec7:dynamic_information}

We consider dynamic, observable information relevant to the \ac{LA} process for a \ac{UE} before a packet transmission, encompassing measurements of radio link conditions, resource availability, and the \ac{UE}'s state. As \ac{LA} operates on a sub-millisecond timescale, accounting for information aging is crucial to accurately predict the optimal \ac{MCS} index. We thus apply either a soft or hard aging model to different information types, as detailed below.

\textbf{Packet Transmission Counter}: Since retransmissions can take advantage of previously received copies of the packet to improve the decoding probability, we include a packet transmission counter $n$ (see~\Cref{eqn:reward}) among the state features. Differentiating between the initial transmission and subsequent retransmissions allows to learn policies that effectively optimize \ac{LA} parameters in multiple transmission attempts of the same packet.

\textbf{\ac{UE}-Specific Measurements}: The \ac{3GPP} \ac{NR} system supports various signals and measurements for channel condition monitoring. We exclusively consider measurements available before packet transmission, including \ac{RSRP} and \ac{CSI} reports. \ac{RSRP} measurements serve as a proxy for pathloss, reflecting macro channel conditions like shadowing and \ac{UE}-cell distance. \ac{CSI} reports, comprising a \ac{CQI} value and recommended transmission rank, capture dynamic channel conditions. While \ac{CSI} reports adhere to technical specifications facilitating (intervendor) interoperability, hardware and algorithm differences among \acp{UE}—made by different manufacturers—may yield variations in reported \ac{CSI} under identical channel conditions.

\textbf{Historical CSI}: Instead of relying solely on the most recent \ac{CSI} report, we consider a sequence of $K_{CSI}\geq 1$ historical \acp{CQI} and rank values, along with a rank-weighted average \ac{CQI}. To account for information aging, we define a time window $W$, setting \ac{CQI} and rank values to $-1$ for reports older than $W$. The window $W$ can correspond to the channel coherence time, allowing models to prioritize recent channel information when determining \ac{LA} parameters. 

\textbf{\ac{UE} Buffer Status:} The \ac{UE} buffer status indicates the amount of data pending transmission. Instead of using the raw buffer size $B_{t}\in\mathbb{R}_{+}$ at time $t\in\mathbb{N}$, we normalize $B_t$ by the maximum transport block size $TBS_{\max}$ allowed by the system:
\begin{align} \label{eq:buffer_norm}
    b_t = \min \left\{ N_{TTI},\; \frac{B_t}{TBS_{\max}} \right\}.
\end{align}
This expression represents the buffer size as the number of transmissions required to clear the buffer under ideal conditions (i.e., most favorable channel and resource availability), with $N_{TTI}$ serving as an upper bound. This normalization renders the buffer state representation independent of system bandwidth and modulation order, enhancing reusability across heterogeneous \acp{RAN}.

\textbf{Expected Spectral Efficiency:} We also incorporate into the \ac{LA} state an estimate of \ac{SE} derived from recent historical \ac{CSI} reports. For any given \ac{UE} $i$, we maintain an average spectral efficiency, defined by
\begin{align}
SE_i = \frac{\sum_{k=1}^{K_{CSI}} SE_k^{(i)} \cdot rank_k^{(i)}}{\sum_{k=1}^{K_{CSI}}  rank_k^{(i)}},
\end{align}
where $SE_k^{(i)}$ and $rank_k^{(i)}$ denote the spectral efficiency and transmission rank associated with the $k$-th historical \ac{CSI} report for \ac{UE} $i$. Specifically, $SE_k^{(i)}$ corresponds to the nominal spectral efficiency associated with the $k$-th reported \ac{CQI}, as defined in the \ac{3GPP} \ac{5G} \ac{NR} specifications~\cite{3GPP38214}.

\textbf{\ac{HARQ} Feedback:} \ac{LA} algorithms traditionally rely on \ac{HARQ} feedback (ACK/NACK) from \acp{UE} to refine link quality estimates. We consider a sequence of $K_{H}$ historical \ac{HARQ} values, $\{h_{k}^{(t_k)}\}_{k=1}^{K_{H}}$, where the superscript $t_k$ denotes the time when the $k$-th report was received. We define $h_{k}^{(t_k)} = 1$ for ACK and $h_{k}^{(t_k)} = -1$ for NACK.

However, the relevance of HARQ feedback diminishes with age—particularly in bursty or ultra-reliable low-latency (URLLC) traffic, or under high cell load, where consecutive transmissions to the same \ac{UE} may occur with long temporal gaps (e.g., exceeding the channel coherence time). In such cases, outdated HARQ reports may no longer reflect the current link state.

To account for this, we apply a soft-aging model that linearly scales $h_{k}^{(t_k)}$ based on its age $t - t_k$ and a maximum aging window $W$:
\begin{equation}
	\tilde{h}_{k}^{(t_k)} =
	\begin{cases}
		h_{k}^{(t_k)}\left(1 - \frac{\max\{0,\; t - t_k\}}{W} \right) & \text{if } t - t_k \leq W \\\\
		0 & \text{otherwise}
	\end{cases}
	\label{eqn:AgedHARQ}
\end{equation}
The resulting value $\tilde{h}_{k}^{(t_k)} \in [-1, 1]$ reflects the aged contribution of each HARQ report, gradually decaying toward zero as its age approaches $W$, thereby reducing its influence on \ac{LA} decisions when it becomes stale.

\textbf{Historical \ac{LA} Parameters}: Historical \ac{LA} parameters (MCS index, rank) from previous transmissions also aid state reconstruction. We include $K_{LA}$ historical MCS indices $\{m_{k}^{(t_k)}\}_{k=1}^{K_{LA}}$ and rank values $\{l_{k}^{(t_k)}\}_{k=1}^{K_{LA}}$. To account for aging, we employ a hard aging model, assigning a value of $-1$ for historical data older than maximum age $W$:
\begin{equation}
\tilde{m}_{k}^{(t_k)} =
\begin{cases}
m_{k}^{(t_k)} & \text{if} \; t-t_k\leq W\\
-1 & \text{otherwise,}
\end{cases}\label{eqn:AgedMCS}
\end{equation}
and
\begin{equation}
\tilde{l}_{k}^{(t_k)} =
\begin{cases}
l_{k}^{(t_k)} & \text{if} \; t-t_k\leq W\\
-1 & \text{otherwise.}
\end{cases}\label{eqn:AgedRANK}
\end{equation}


\section{Experimental Evaluations}\label{sec:numerical_examples}


\subsection{Training Setup}\label{sec:training_setup}


\subsubsection{Distributed Training Architecture}

Our training system, illustrated in~\Cref{fig:APEX}, utilizes a distributed architecture comprising a single learner, 40 actors, and a shared replay memory composed of several independent shards. These components are co-located on a shared compute node, while the simulation environments are hosted on separate compute nodes to facilitate a distributed and scalable experience collection. To maximize throughput, the system operates asynchronously, allowing actors and the learner to work independently without global synchronization.

The replay memory is partitioned into four independently prioritized shards, each maintaining its own priority queue. Each shard has a capacity of 4 million transitions, enabling long-term experience storage and sampling diversity. Actors write transitions asynchronously to specific shards in batches, reducing communication overhead and supporting scalable data ingestion, similar to~\cite{KCJ+:22}.

\begin{figure}
    \centering
    \includegraphics[width=0.7\columnwidth]{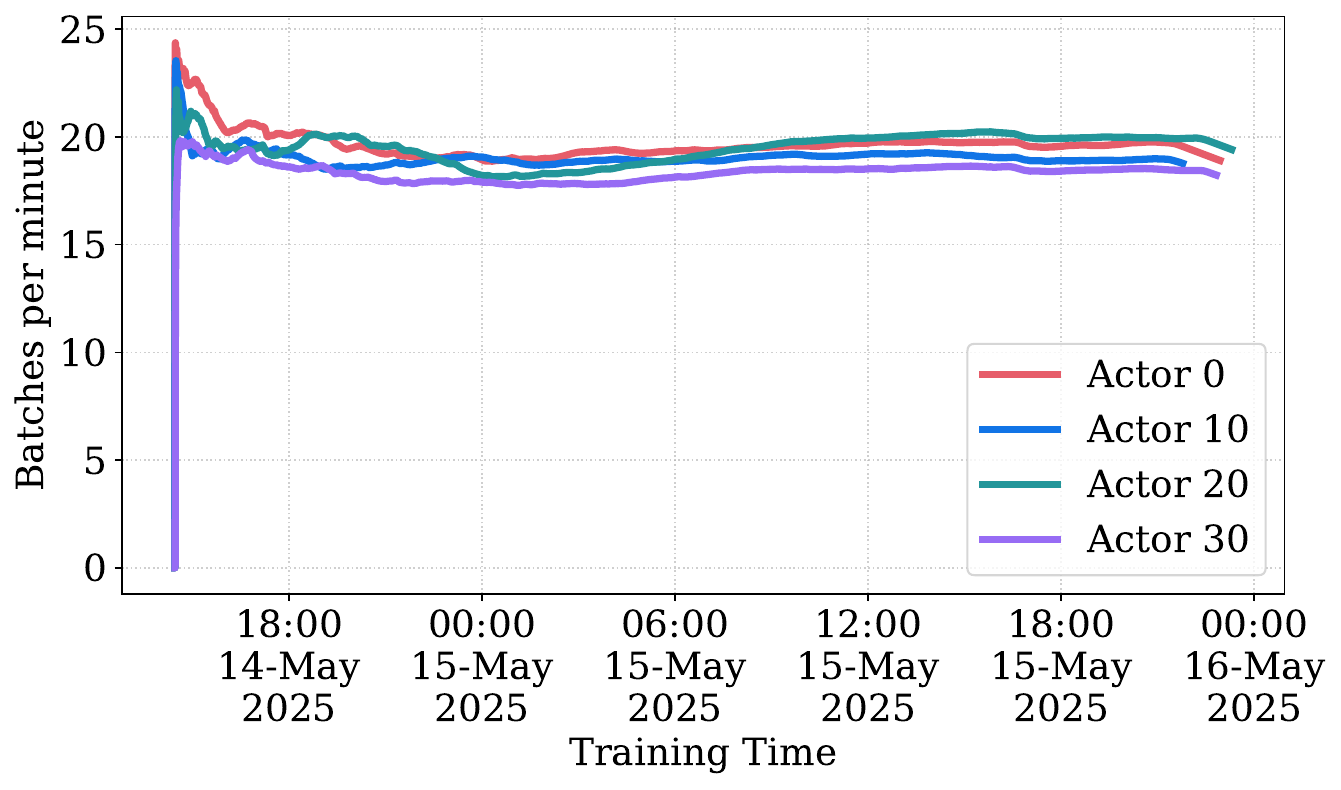}
    \caption{\textbf{Per-actor replay-memory ingestion rates during large-scale distributed training.} Each curve shows the number of 500-sample batches written per minute by four representative actors (IDs 0, 10, 20, and 30) to their assigned shards over a 30-hour run. After an initial ramp-up and shard initialization jitter (14 May, 17:00--19:00 UTC), all actors converge to a steady rate of 18--20 batches/minute. The uniformity of these curves confirms consistent ingestion throughput and balanced load across the actor population.}
    \label{fig:0_a}
\end{figure}

As shown in~\Cref{fig:0_a}, each actor steadily writes batches to its assigned shard at a rate of approximately 18–20 batches per minute after an initial ramp-up period. This time series, covering a representative 30-hour training run, displays data from four actors (IDs 0, 10, 20, and 30) and highlights both the system’s ability to reach a stable ingestion regime and the even distribution of load across actors and shards. The close alignment in write rates across disparate actor indices provides strong empirical evidence of the scalability and load-balancing capabilities of our distributed experience collection infrastructure.

Actors run local replicas of the policy network and interact concurrently with 14 environment instances (\Cref{app:compute_resources} provides additional computation and deployment details). Similar to~\cite{Horgan:2018}, each actor $i \in \{0, \cdots, 39\}$ executes an $\epsilon_i$-greedy policy, defined as $\epsilon_i = \epsilon^{1 + \frac{i}{N-1} \alpha}$, where $\epsilon$ is set to 0.4 and $\alpha$ to 8.5. This exploration policy remains fixed throughout training. On average, each actor generates approximately 132 transitions per second. Incoming transitions are buffered locally before being dispatched asynchronously to a replay memory in batches of 500 transitions. Prior to training, a warm-up phase collects 45{,}000 transitions—without gradient updates—to populate the replay memory with a diverse initial dataset.

The learner is responsible for computing gradients and updating model parameters. It prefetches up to 16 batches of 512 transitions each from the replay memory shards. This prefetching ensures a continuous supply of data for gradient computation. On average, the learner processes 100 batches per second, corresponding to approximately 51{,}200 transitions per second. Model parameters are updated and broadcast asynchronously to all actors every 512 learner steps (roughly every \qty{5}{\second}), allowing actors to operate with relatively fresh policies while maintaining a decoupled learning and acting process.

The data flow proceeds as follows: actors collect environment transitions, locally buffer them, and push them to replay memory shards. The learner samples from these shards, performs gradient updates, and propagates the updated parameters back to the actors.

\begin{table}[t]
	\caption{RAN environment simulation parameters. Bold values are default parameters used in testing scenarios (which randomize over seeds only).} 
	\centering 
	\begin{tabular}{l l l} 
		\toprule[1pt]\midrule[0.3pt]
		\textbf{Parameter} & \textbf{Value range} & \textbf{Description} \\ [0.5ex]
		\midrule
		Duplexing type & TDD & Fixed\\
		Carrier frequency & 3.5\,\si{\giga\hertz} & Fixed \\
		Deployment type & 1-site 3-sector & \\
		Site type & \{MIMO, mMIMO\} & Randomized\\
		Antenna array & 1x2x2 MIMO (4) & Fixed  \\
		&8x4x2 mMIMO (64) & Fixed \\
		Cell radius & \{\textbf{166}, 300, 600, 900, 1200\}\,\si{\meter} & Randomized \\
		Bandwidth & \{\textbf{20}, 40, 50, 80, 100\}\,\si{\mega\hertz} & Randomized\\
		Number of sub-bands & \{\textbf{51}, 106, 133, 217, 273\} & Randomized \\
		DL TX power & \{\textbf{20}, 40, 50, 80, 100\}\,\si{\watt}  & Randomized \\
		UE antennas & \{2, \textbf{4}\} & Randomized \\
		Max transm. rank & \{2, \textbf{4}\} & Randomized \\
		Max num. transm. & 5 & Fixed \\
		UE traffic type & \{FB, eMBB\} & Randomized \\
		Number FB UEs  & \{1, 5, \textbf{10}\} & Randomized \\
		Number MBB UEs & \{5, \textbf{10}, 30, 50, 100\} & Randomized\\
		Speed UE FB & \{\textbf{0.67}, 10, 15, 30\}\,\si{\metre\per\second} & Randomized \\
		Speed UE eMBB & \{\textbf{0.67}, 1.5, 3\}\,\si{\metre\per\second} & Randomized\\
		UE receiver types & \{\textbf{type0}, type1, type2, type3\} & Randomized \\
		Indoor probability & \{0.2, 0.4, \textbf{0.8}\} & Randomized\\
		\midrule[0.3pt]
	\end{tabular}\label{table:sim_params}
\end{table}


\subsubsection{Training Environment Diversification}

We train the \ac{RL} policy to generalize in the \ac{RAN} environment by exposing it to a wide range of deployment scenarios, radio conditions, traffic patterns, interference levels, and load distributions, as discussed in~\Cref{sec3b:enabler_diversification}. While such diversity is naturally present in live \ac{RAN} systems—where training data can be generated across network nodes—we use domain randomization to recreate comparable conditions within network simulators.

\begin{table*}[t!]
	\caption{Benchmark network configuration scenarios used for testing.}
	\centering
    \resizebox{\textwidth}{!}{
	\begin{tabular}{c l l l l}
		\toprule[1pt]\midrule[0.3pt]
		\textbf{Name} & \textbf{Type} & \textbf{Description} & \textbf{Traffic} & \textbf{User Types} \\ [0.5ex]
		\midrule
        B1 & Single-cell single UE & 1-site 1-cell MIMO, 1 outdoor UE at different locations & FB & Homogeneous (type0) \\
		B2 & Stable interference & 1-site 3-cell (a. MIMO, b. mMIMO), 80\% indoor UEs & FB & Homogeneous (type0) \\
		B3 & Dynamic interference & 1-site 3-cell (mMIMO), 80\% indoor \acp{UE} & eMBB & Homogeneous (type0) \\
		B4 & High mobility & 1-site 3-cell (mMIMO), 0\% indoor \acp{UE} & FB & Homogeneous (type0) \\
		B5 & Realistic traffic & 3-site 9-cell (mMIMO), 80\% indoor \acp{UE} & FB and eMBB & Homogeneous (type0) \\
		\midrule[0.3pt]
	\end{tabular}
    }\label{table:benchmarks}
\end{table*}

Each training scenario is generated using an event-driven, \ac{5G}-compliant system simulator that operates in \ac{TDD} mode on a \qty{3.5}{\giga\hertz} carrier frequency, with \ac{SU-MIMO} transmission and a physical layer model based on numerology $\mu=0$ as specified in \ac{3GPP} \ac{NR} standard 38.211~\cite{3GPP38211}.

A training scenario consists of three trisectorial radio sites, each randomly configured as either \ac{MIMO} or \ac{mMIMO}. Site configurations are randomized with respect to cell radius, system bandwidth, and downlink transmit power, using parameters sampled from~\Cref{table:sim_params}. Further randomization is applied to cell load, traffic type, indoor/outdoor \ac{UE} distribution, and \ac{UE} receiver types. A \ac{UE} generation process randomly selects the number of \acp{UE} with \ac{FB} and \ac{eMBB} traffic to be deployed, drawing from a predefined set of indoor/outdoor probabilities listed in~\Cref{table:sim_params}.

\acp{UE} with \ac{FB} traffic always have data available for scheduling. Consequently, simulations containing \emph{only} \ac{FB} traffic exhibit a stable interference environment, as the entire frequency band is fully utilized in every downlink slot. In contrast, \ac{eMBB} \acp{UE} generate bursty traffic, leading to a more dynamic and time-varying interference pattern. Each \ac{eMBB} \ac{UE} generates traffic according to a Poisson-based packet arrival process with parameters inferred from field data. The dataset spans multiple user sessions and captures a wide range of application behaviors, ranging from short, bursty transmissions to sustained data transfers. These behaviors, in turn, influence both packet inter-arrival times and packet size statistics. Payload sizes in the model range from approximately \qty{20}{\kilo\byte} to \qty{1000}{\kilo\byte}, reflecting the diversity of application types.

Each \ac{UE} configuration is further randomized by the number of antennas, mobility speed, and receiver type. The latter accounts for variations in channel estimation accuracy among different \ac{UE} chipsets under identical channel conditions.

For hyperparameter tuning, summarized in~\Cref{app:hyperparams}, we train \ac{RL} policies on 40{,}000 training scenarios generated from 160 random seeds and 250 randomized simulation parameter configurations, as defined in~\Cref{table:sim_params}. Each scenario simulates five seconds (equivalent to 5{,}000 \acp{TTI}), producing approximately 16{,}000 training samples on average and producing a total of approximately 650 million samples across the entire sweep.

\Cref{sec8c:general_results} presents the general performance evaluation employing a larger dataset of 240{,}000 training scenarios generated from the same 160 seeds but with 1{,}500 randomized parameter configurations, also drawn from~\Cref{table:sim_params}. This setup produces approximately 3 billion training samples.

\begin{figure*}[t]
    \centering
    \subfloat[Cell throughput relative gains across benchmark scenarios.]{ \includegraphics[width=0.49\textwidth]{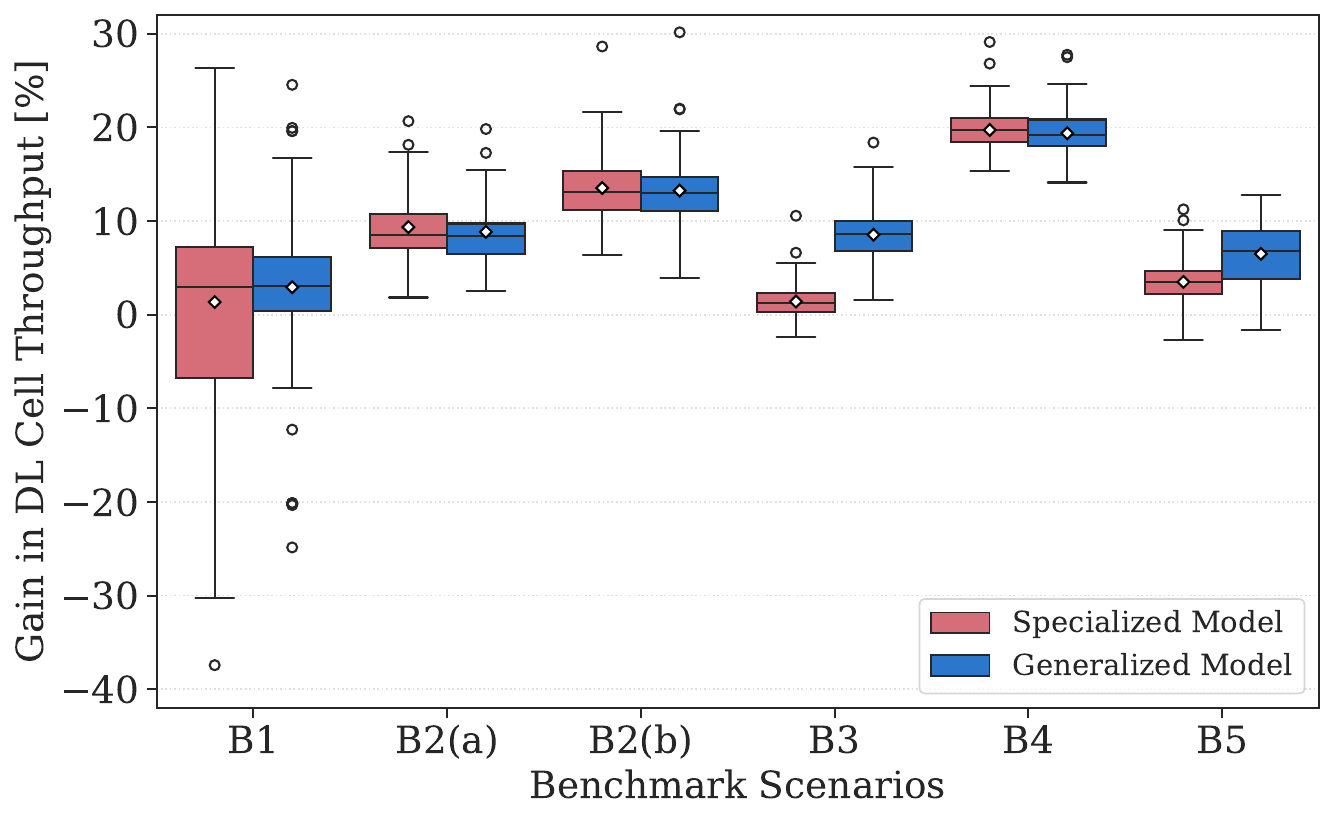}
    \label{fig:gen_res_cell_throughput}}%
    \subfloat[Spectral efficiency relative gains across benchmark scenarios.]{\includegraphics[width=0.49\textwidth]{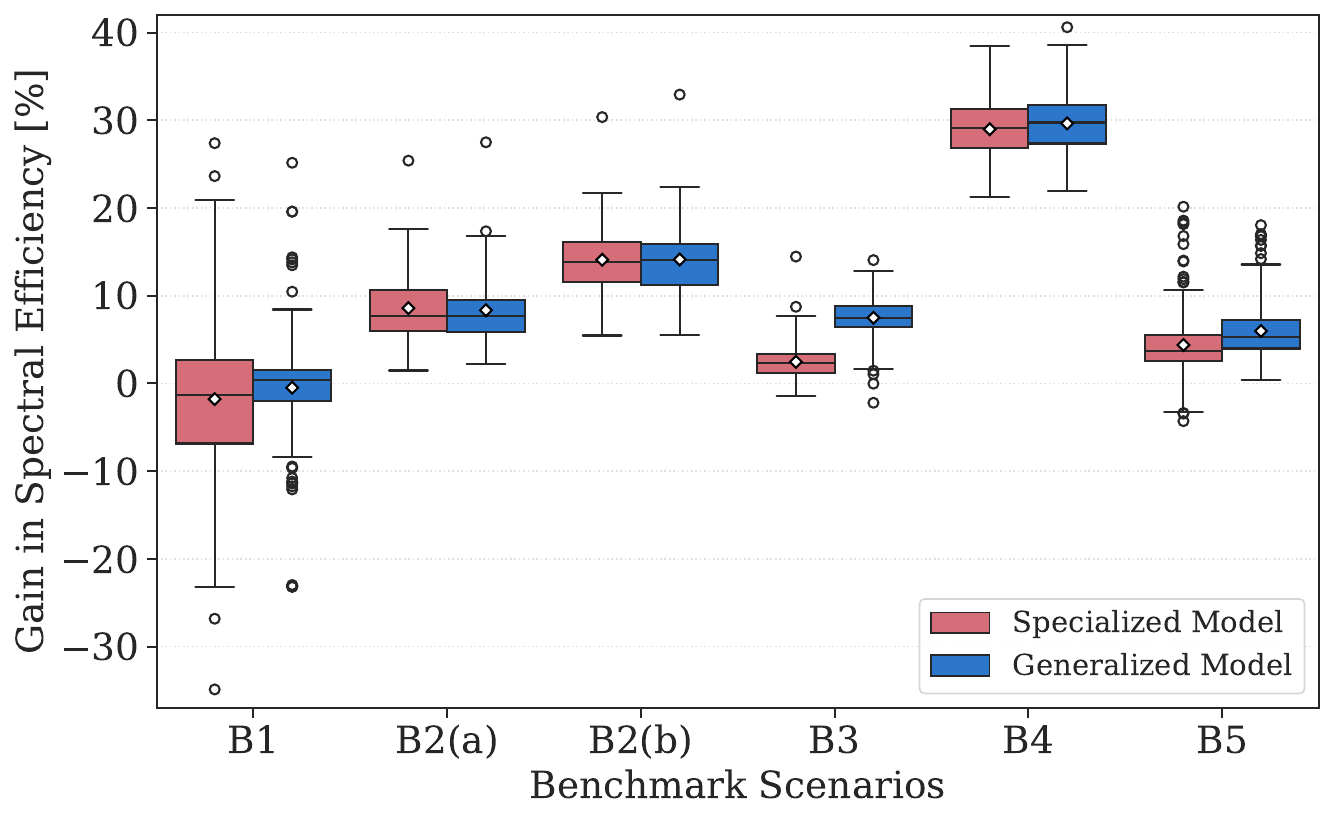}
    \label{fig:gen_res_spectral_efficiency}}
    \caption{Performance comparison of specialized and generalized \ac{RL}-based \ac{LA} policies across five benchmark scenarios (B1--B5). The boxplots illustrate relative gain or loss (\%) over a state-of-the-art baseline for two \ac{RL}-based \ac{LA} policies, both employing the same neural network architectures: a six-layer multilayer perceptron with 256 neurons per layer. The policies differ only by training procedure: a specialized policy trained exclusively for full-buffer (FB) conditions (red), and a generalized policy trained across varied network conditions (blue). Results are presented for five benchmark scenarios (B1--B5), showing the distribution of relative performance gains in: (a) downlink cell throughput, and (b) spectral efficiency. Each boxplot displays the median (diamond), interquartile ranges (box), whiskers extending to $2\times$IQR, and outliers (circles).}
    \label{fig:general_results}
\end{figure*}


\subsection{Test Setup}

For testing purposes, we define five benchmark scenarios, summarized in~\Cref{table:benchmarks}, whose conditions are not directly encountered during training. Benchmark B1 corresponds to a single-cell, single-UE deployment operating in \ac{MIMO} transmission mode. This benchmark is used to assess link-level performance, where the \ac{UE} is placed at predefined distances from the base station, and the channel realization is randomized across multiple simulations per \ac{UE} position. This scenario serves as a challenging outlier for evaluating the link-level performance of \ac{RL}-based policies against the state-of-the-art \ac{OLLA} algorithm used in \ac{5G} networks, which is specifically optimized for such conditions.

Scenarios B2--B5 consist of homogeneous network deployments, either \ac{MIMO} or \ac{mMIMO}, with simulation parameters using the default values indicated in bold in~\Cref{table:sim_params}. Each benchmark targets specific testing conditions: B2 evaluates stable interference under \ac{FB} traffic using either MIMO (B2(a)) or mMIMO (B2(b)) deployments; B3 simulates dynamic interference in \ac{mMIMO} deployments due to \ac{eMBB} traffic; B4 assesses performance for \acp{UE} traveling at high speeds (\qtyrange{100}{200}{\kilo\metre\per\hour}); and B5 evaluates performance under realistic traffic conditions involving a mix of \ac{FB} file transfers and chat traffic types unseen during training.


\subsection{General Results} \label{sec8c:general_results}

To validate our design concepts for \ac{RL} generalization in \ac{RAN} applications, we compare the performance of two \ac{LA} policies trained using the same model architecture and hyperparameters but under different data regimes. The first policy is trained to generalize using randomized network scenarios, while the second is \textit{specialized} for \ac{FB} traffic conditions by using an equal amount of simulated networks based on benchmarks B2(a), B2(b), and B4, with randomized seeds. In both cases, we use an \ac{MLP} with six hidden layers and 256 neurons per layer. Inputs are described in~\Cref{sec:7d}, and the output layer comprises 28 nodes, each representing the Q-value associated with an \ac{MCS} index compliant with \ac{5G} specifications (Table 5.1.3.1-2 in \ac{3GPP} TS 38.214~\cite{3GPP38214}). Both policies are trained using a robustness weight $\alpha=0.5$ (see~\Cref{eqn:reward}). 

We compare both policies with a state-of-the-art \ac{OLLA} design that targets a $10\%$ \ac{BLER}, representing the most common approach adopted in real-world \ac{5G} \ac{NR} systems.

\Cref{fig:general_results} summarizes the test results obtained across the five benchmarks described in~\Cref{table:benchmarks}, reporting statistics of gains and losses in average cell throughput (cf.~\Cref{fig:gen_res_cell_throughput}) and spectral efficiency (cf.~\Cref{fig:gen_res_spectral_efficiency}) relative to the baseline. 

First, we observe that the generalized \ac{RL} policy significantly outperforms the 5G baseline across all benchmarks, achieving gains in average cell throughput and spectral efficiency close to or exceeding 10\% in \ac{FB} \ac{MIMO} and \ac{mMIMO} scenarios (i.e., B2(a)-(b)), and approximately 20\% in high-mobility conditions (i.e., B4).

Second, under \ac{FB} traffic conditions (i.e., benchmarks B2 and B4), the generalized policy performs on par with the specialized policy, while significantly outperforming it in B3 (\ac{eMBB} traffic) and B5 (mixed traffic), with gains of approximately 4× and 2×, respectively. In benchmark B1—analyzed in more detail in the following subsection—the generalized model also achieves higher gains with lower variance in both throughput and spectral efficiency.

These results suggest that \ac{RL} algorithms trained for generalization using the enablers proposed in this paper can outperform both heuristic baselines and models optimized for narrow, scenario-specific conditions. Unlike specialized models, which perform well only under conditions seen during training and often lack robustness, the generalized approach proposed here provides the adaptability needed for scalable \ac{RL} deployment in realistic networks, reducing the need for frequent retraining.

Additional evaluations on hyperparameter tuning, including model architecture, layer normalization, and loss functions, are presented in~\Cref{app:network_architecture}.

\begin{figure*}[th!]
    \centering
    \subfloat[Average throughput gains in benchmark B1.]{ \includegraphics[width=0.49\textwidth]{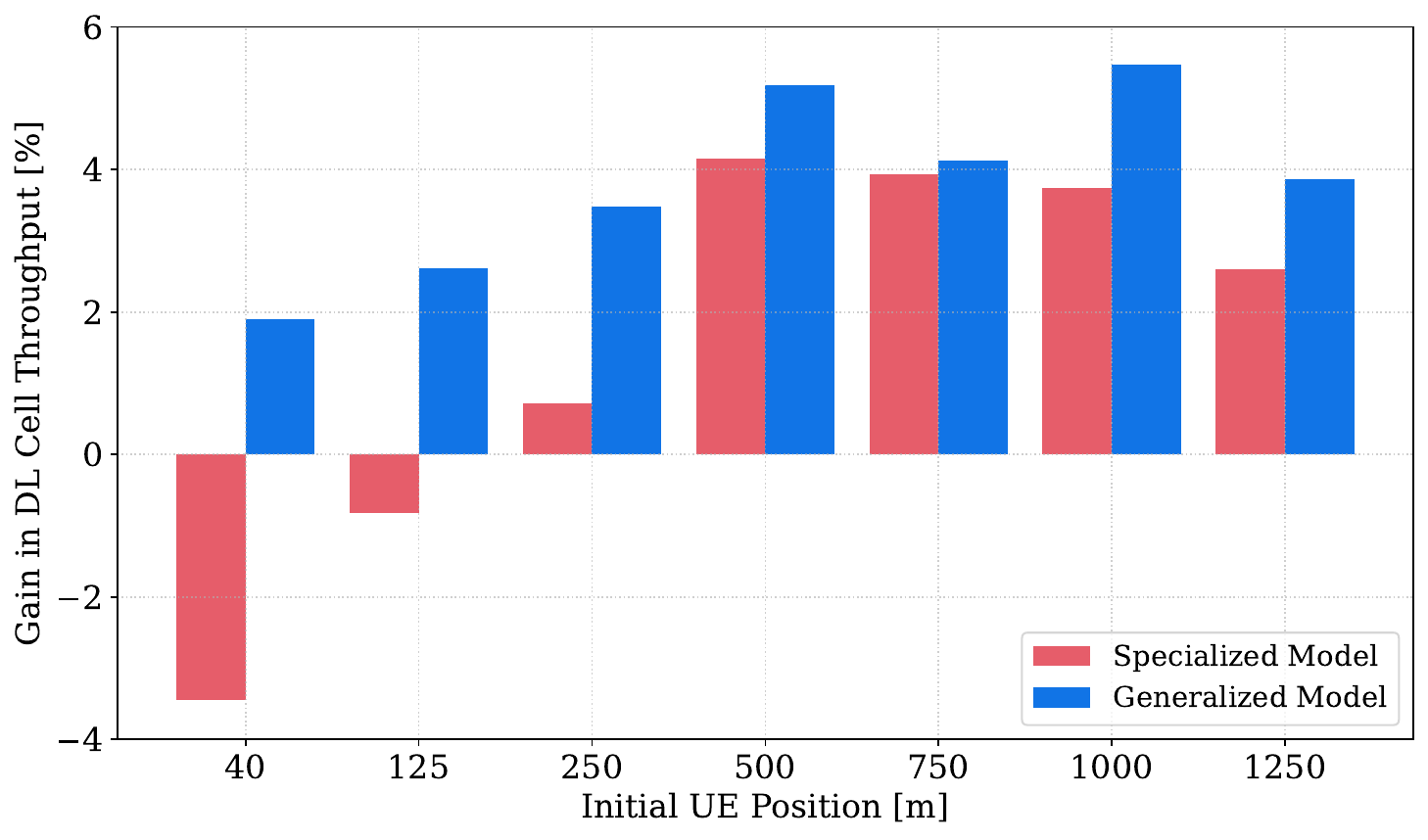}
    \label{fig:SUCU_throughput_gains}}%
    \subfloat[Average \ac{MCS} index selected in benchmark B1.]{\includegraphics[width=0.49\textwidth]{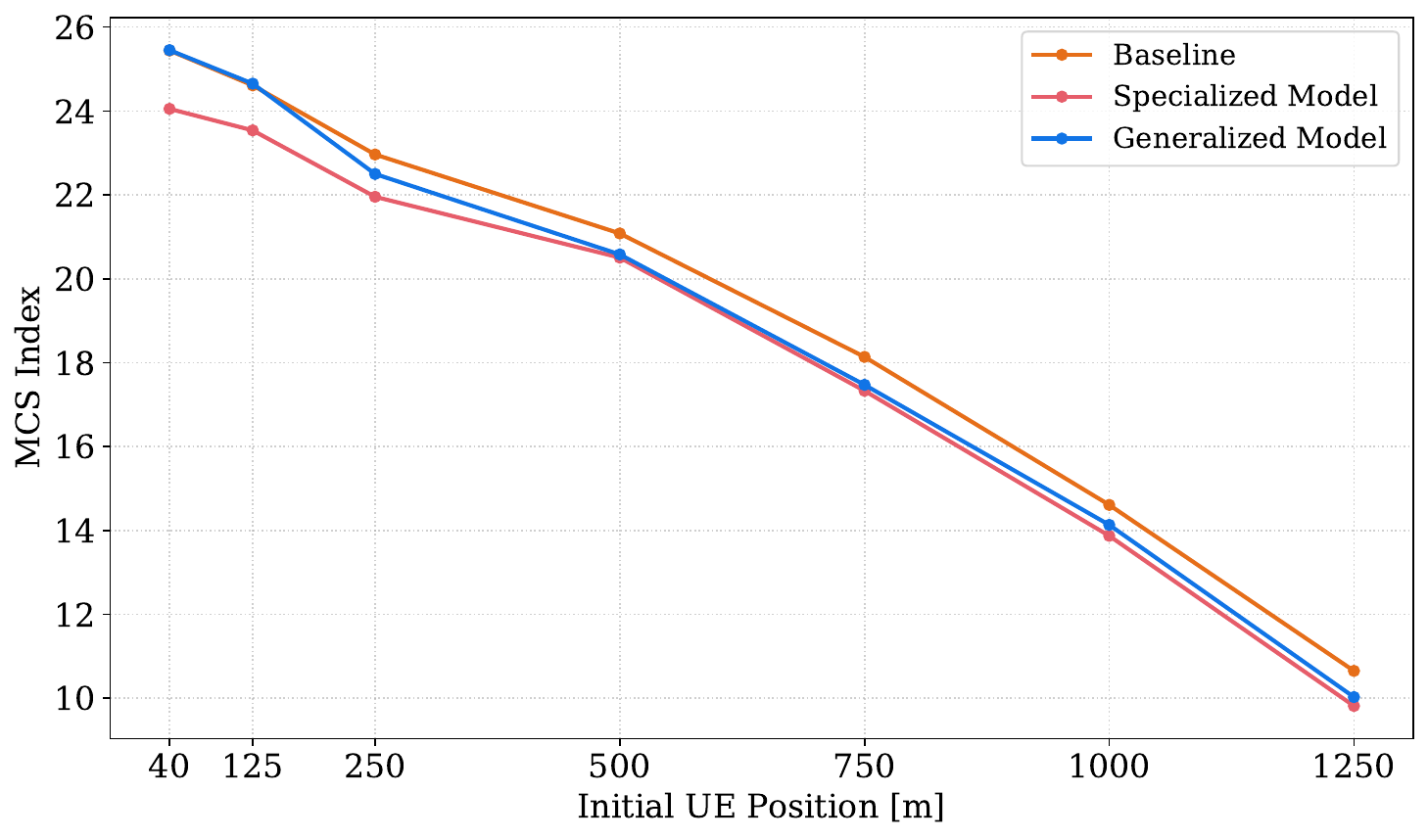}
	\label{fig:SCSU_MCS_distribution}}
    \caption{Performance analysis in single-cell single-UE benchmark B1, with the model tested on several realizations of the scenario with different \ac{UE} positions ranging from 40\,\text{m} to 1250\,\text{m} from the serving cell. Throughput gains are measured with respect to the \ac{OLLA} baseline.}
    \label{fig:SCSU_generalized_model}
\end{figure*}


\subsubsection{Link-Level Performance}

\Cref{fig:SCSU_generalized_model} further examines the performance of both models in the single-cell single-\ac{UE} benchmark B1. This benchmark is relevant for two main reasons: it reflects outlier conditions never encountered during training, and it represents the scenario for which the \ac{OLLA} baseline is optimized. The latter provides reference values for link-level spectral efficiency used, for instance, by \ac{3GPP} to standardize \ac{CQI} and \ac{MCS} tables~\cite{3GPP38214}.  

We evaluate both the generalized and specialized models against the baseline across several initial \ac{UE} locations, ranging from \qty{40}{\meter} to \qty{1250}{\meter} from the serving base station. At each location, we conduct 50 simulations using distinct random seeds for channel and environment realizations. \Cref{fig:SUCU_throughput_gains} shows that the generalized \ac{RL} model consistently outperforms both the baseline and the specialized model at all locations. While average throughput gains range between 2\% and 5.5\%, it is important to note that the \ac{OLLA} baseline nearly reaches the link capacity under B1 conditions, leaving limited room for improvement. 

\Cref{fig:SCSU_MCS_distribution} shows that the generalized \ac{RL} model achieves these gains by selecting, on average, slightly more conservative \ac{MCS} indices at greater distances from the base station. This conservatism reduces the number of retransmissions experienced by the \ac{RL} policy, resulting in modest but consistent throughput gains. In contrast, the specialized model is overly conservative: near the base station, it fails to recognize ideal channel conditions (i.e., low path loss attenuation), selecting instead lower \ac{MCS} indices. This results in fewer information bits being transmitted and thus a performance loss. At other distances, this model also tends to select slightly more conservative \ac{MCS} indices, yielding smaller gains over the baseline.


\subsection{Reward Analysis}

\begin{figure*}[t]
  \centering
  \subfloat[Cell throughput statistics vs.\ $\alpha$.]{%
    \includegraphics[width=0.49\textwidth]{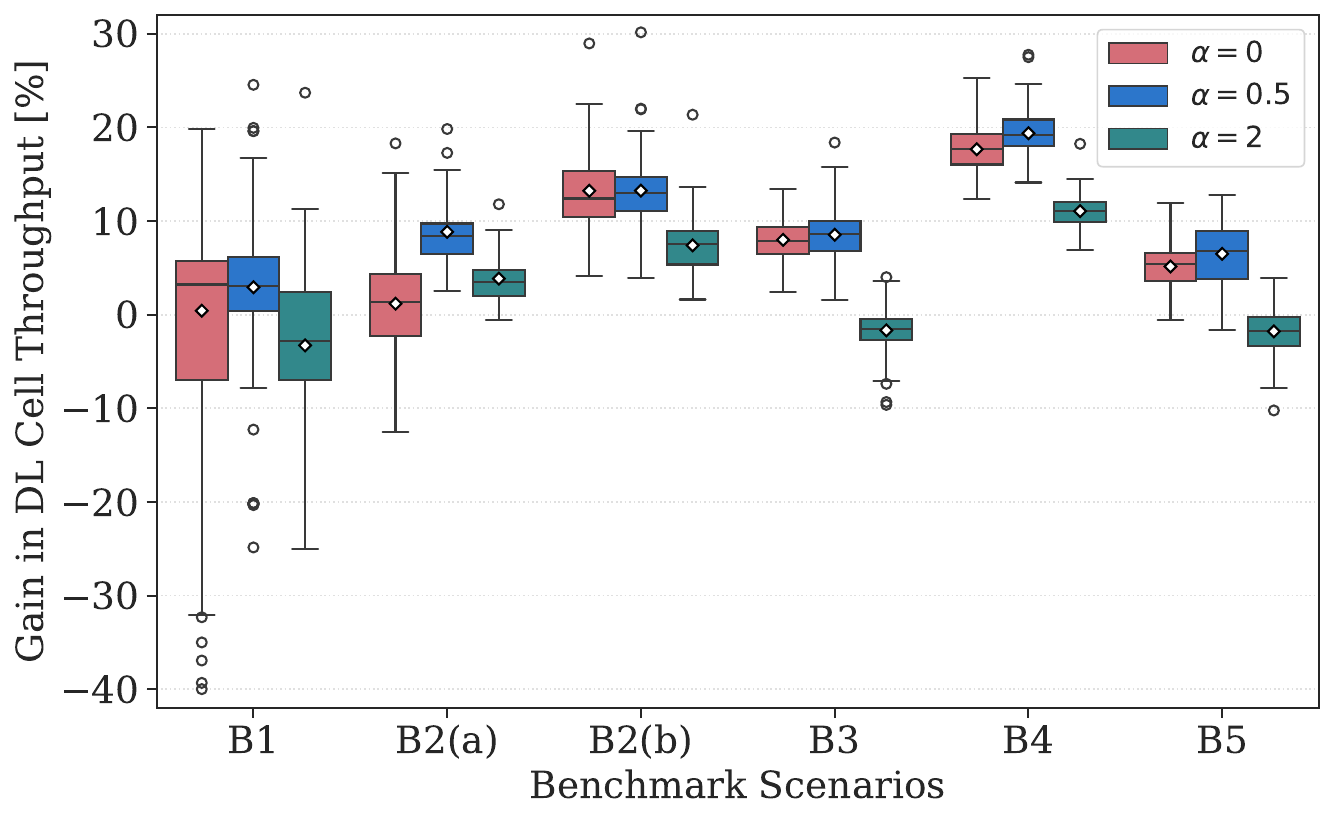}%
    \label{fig:7a_throughput}%
  }%
  \hfill
  \subfloat[Spectral efficiency statistics vs.\ $\alpha$.]{%
    \includegraphics[width=0.49\textwidth]{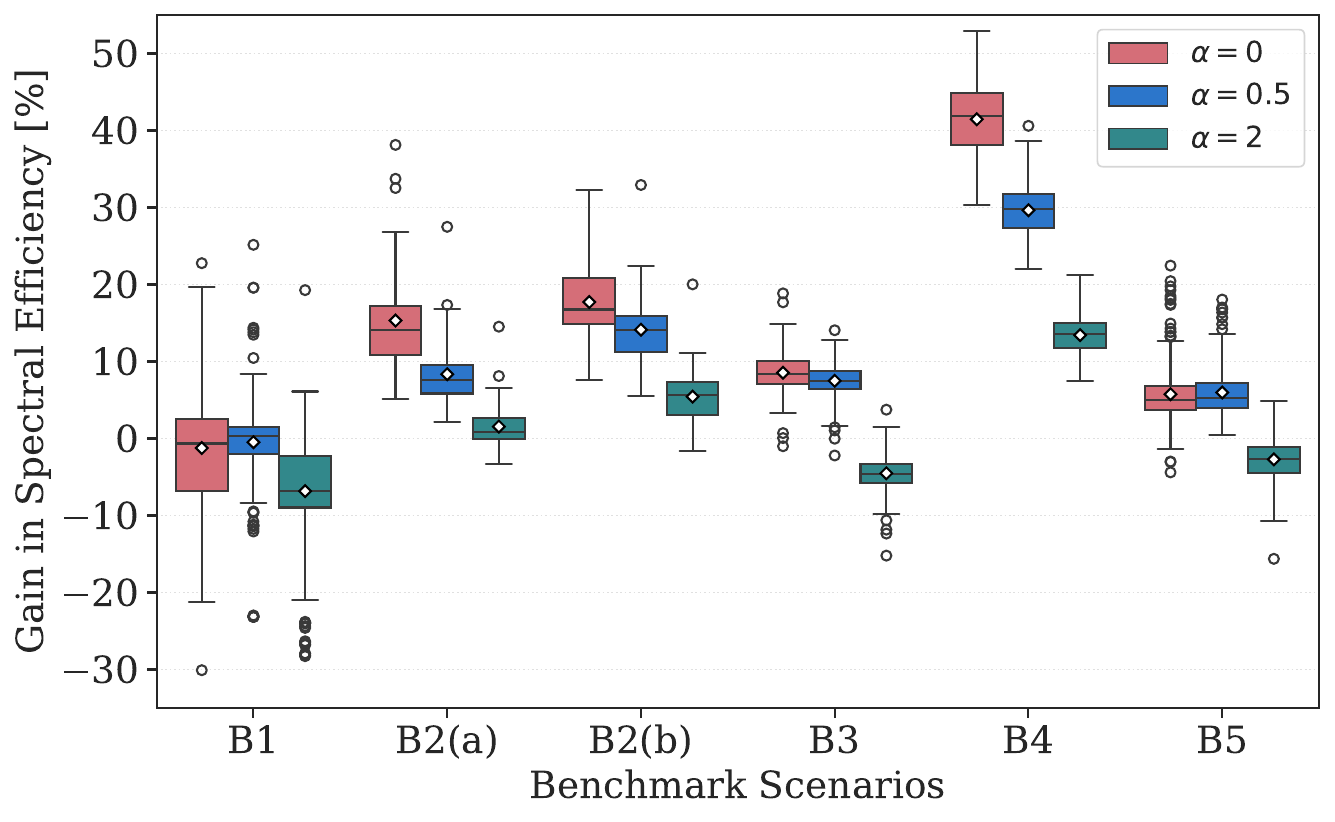}%
    \label{fig:7b_se}%
  }%

  \subfloat[Mean \ac{BLER} vs.\ $\alpha$.]{%
    \includegraphics[width=0.49\textwidth]{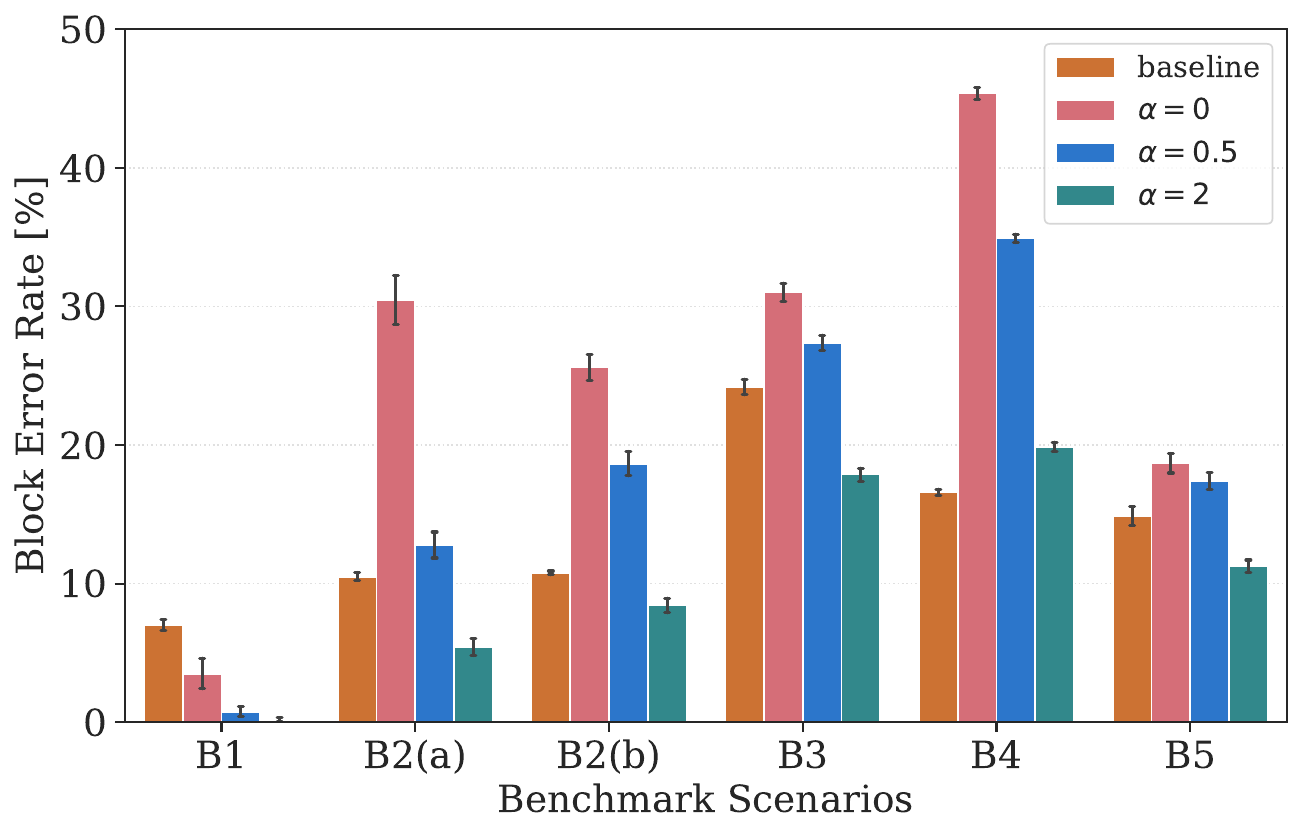}%
    \label{fig:7c_bler}%
  }%
  \hfill
  \subfloat[Mean \ac{MCS} index selected vs.\ $\alpha$.]{%
    \includegraphics[width=0.49\textwidth]{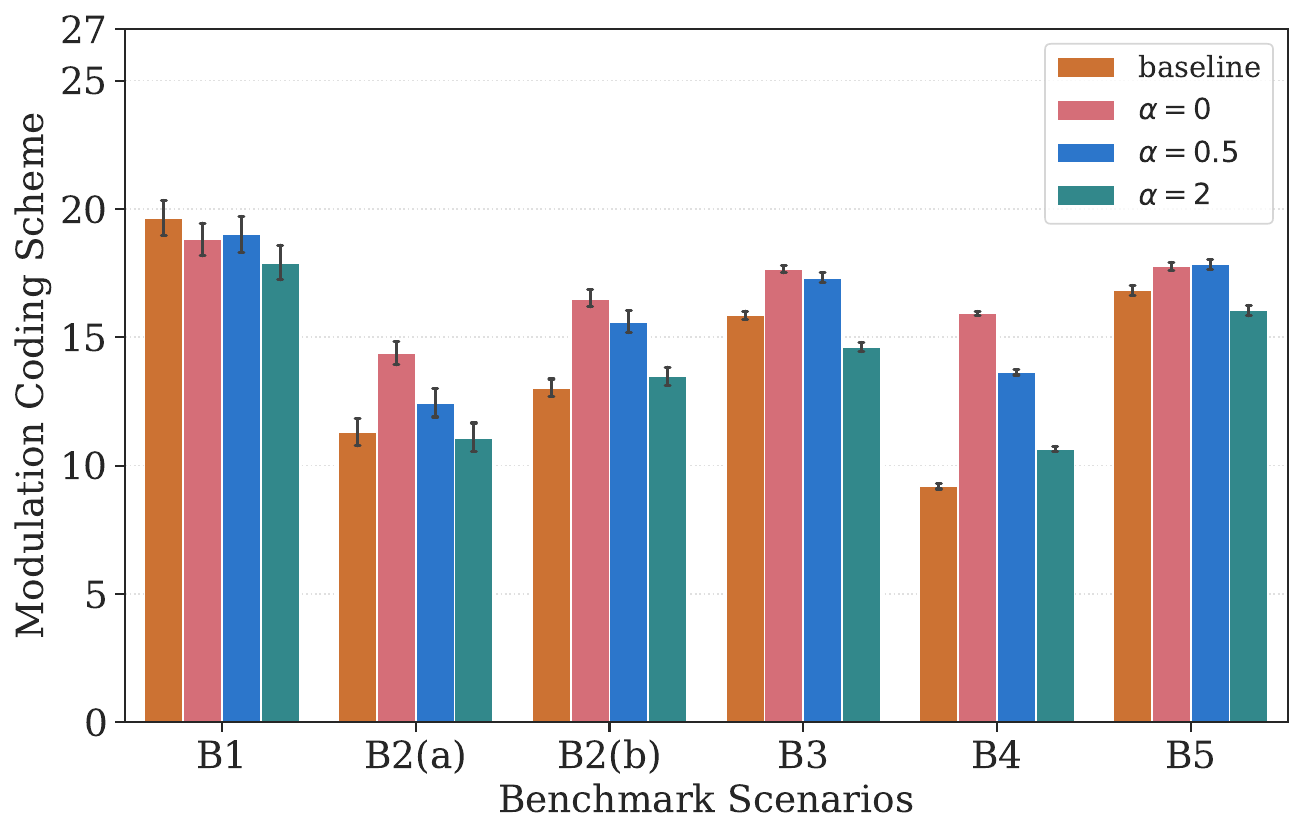}%
    \label{fig:7d_mcs}%
  }%

  \caption{Effect of robustness weight $\alpha$ in the reward function on key performance metrics across five benchmark scenarios (B1--B5). Each column of panels represents to a different performance measure evaluated at three robustness weight levels: no penalty ($\alpha=0$), moderate penalty ($\alpha=0.5$), strong penalty ($\alpha=2$). (a) Distribution of percentage gains in downlink cell throughput relative to a baseline policy, summarized as boxplots from multiple Monte Carlo trials (boxes: interquartile range; whiskers: $2\times\text{IQR}$; circles: outliers; diamonds: means). (b) Distribution of percentage gains in spectral efficiency, using the same boxplot conventions as panel (a). (c) Mean \ac{BLER} with 95\% confidence intervals, demonstrating systematic reductions in error rates with increasing retransmission penalization across scenarios. (d) Mean \ac{MCS} index selected, with 95\% confidence intervals, indicating that higher values of $\alpha$ lead to more conservative (lower-rate) transmission modes.}
  \label{fig:reward_param_comp}
\end{figure*}

\begin{figure*}[th!]
    \centering
    \subfloat[Action distributions for different $\alpha$.]{ \includegraphics[width=0.48\textwidth]{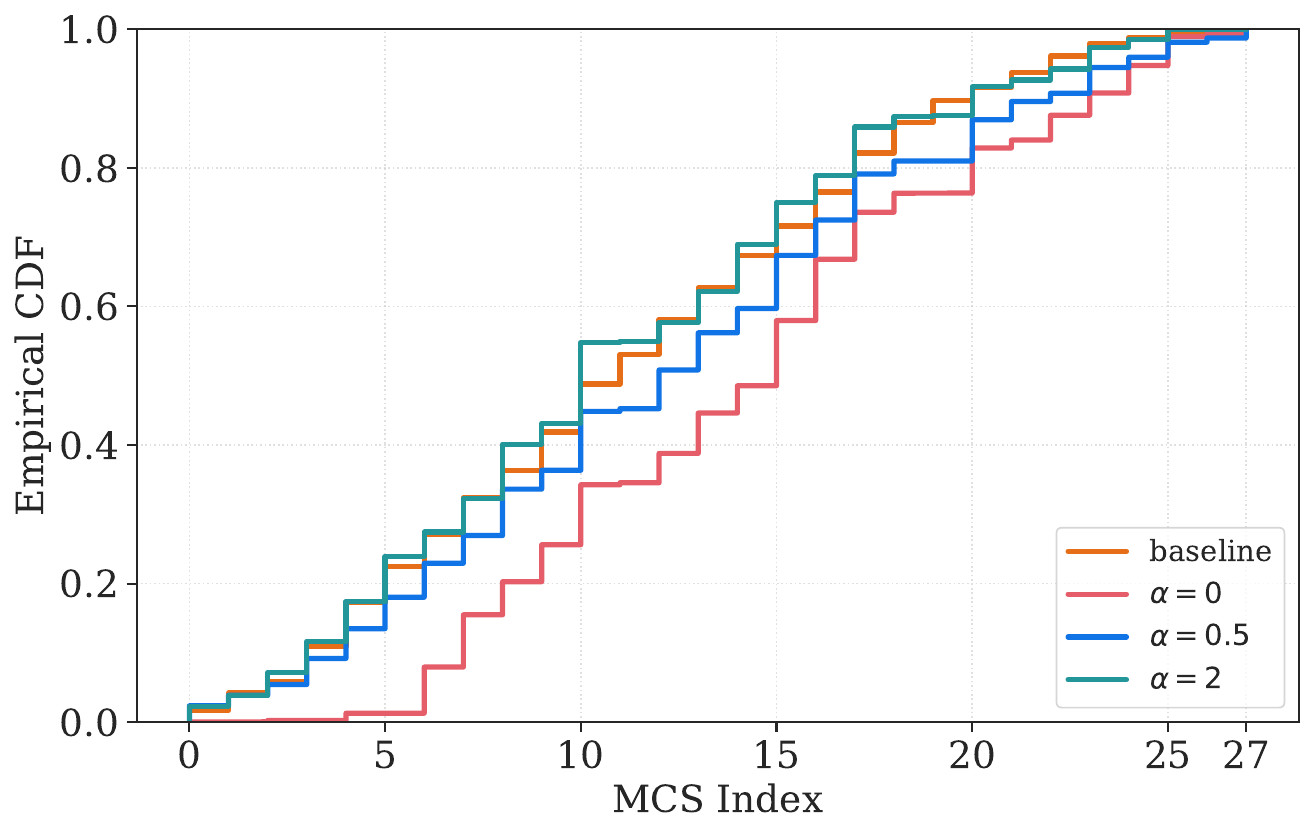}
    \label{fig:8a_mcs_distribution_alpha}}%
    \subfloat[BLER distributions for different $\alpha$.]{\includegraphics[width=0.48\textwidth]{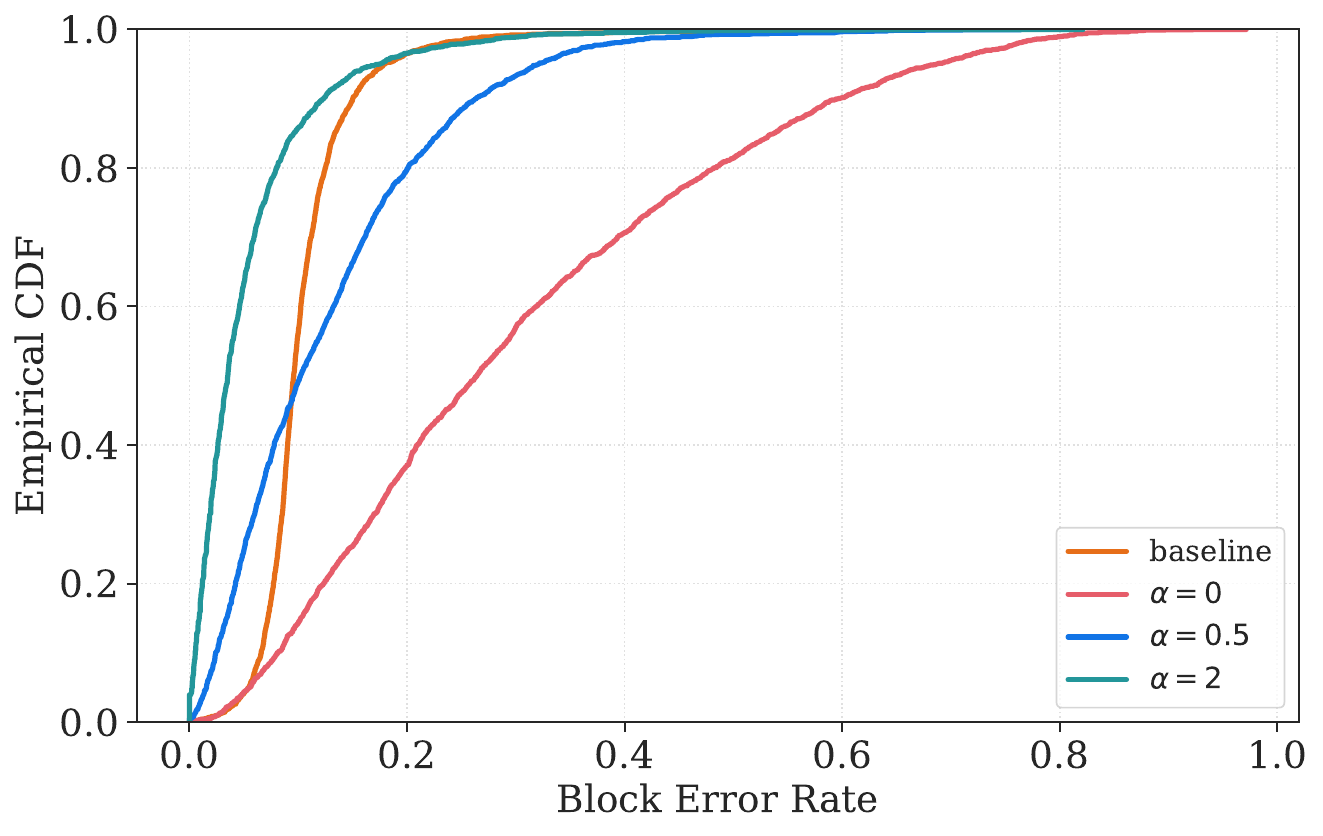}%
	\label{fig:8b_bler_distribution_alpha}}
    \caption{Effect of robustness weight $\alpha$ in the reward function on the learned \ac{RL} policy, illustrated through action distributions (\ac{MCS} index) and \ac{BLER} for three polices evaluated in benchmark scenario B2 from~\Cref{fig:reward_param_comp}. Increasing $\alpha$ from 0 to 2 encourages more conservative \ac{MCS} selection, resulting in smaller data payloads better matched to link capacity, thereby reducing \ac{BLER} distribution.}
    \label{fig:8_performance_vs_alpha_B2}
\end{figure*}

Unlike the \ac{OLLA} baseline, which operates under a 10\% \ac{BLER} target that can be reliably achieved with full-buffer traffic (e.g., benchmark~B2), the proposed \ac{RL}-based design has no explicit access to or control over \ac{BLER} during training or inference. \Cref{fig:reward_param_comp,fig:8_performance_vs_alpha_B2}, together with the reward definition in~\Cref{eqn:reward}, illustrate the \emph{dual role} of the robustness weight $\alpha$ in the \ac{RL}-based \ac{LA} design, enabling the learning of policies that realize different tradeoffs among throughput, spectral efficiency, and \ac{BLER}.

Specifically, we trained three generalized policies using the same model architecture, hyperparameters, and training procedure as in the previous section, but with different settings of $\alpha$: 0, 0.5, and 2. \Cref{fig:reward_param_comp} shows how $\alpha$ affects the performance achieved by each policy when tested in the five benchmarks, presenting statistics of gains/losses in throughput (see~\Cref{fig:7a_throughput}) and spectral efficiency (see~\Cref{fig:7b_se}) relative to the \ac{OLLA} baseline, as well as the average \ac{BLER} (see~\Cref{fig:7c_bler}) and average \ac{MCS} index selected (see~\Cref{fig:7d_mcs}), with 95\% confidence intervals. Moreover, \Cref{fig:8a_mcs_distribution_alpha,fig:8b_bler_distribution_alpha} further examine how $\alpha$ influences the learned policy by showing the \ac{CDF} of the selected \ac{MCS} index (i.e., the action distribution) and the corresponding \ac{BLER}, respectively, for benchmark B2.

\textbf{Setting $\boldsymbol{\alpha = 0}$} deactivates the penalty term for failed transmissions (i.e., NACK receptions). This results in a policy that encourages the selection of more aggressive (i.e., higher) \ac{MCS} indices, as shown in~\Cref{fig:8a_mcs_distribution_alpha}, thereby increasing the raw symbol rate. Consequently, the policy learned with $\alpha = 0$ tends to maximize spectral efficiency, as illustrated in~\Cref{fig:7b_se} across all benchmarks, but at the expense of lower throughput, as depicted in~\Cref{fig:7a_throughput}. Specifically, the policy exploits retransmissions to deliver more information bits than the channel can support in a single transmission, resulting in high \ac{BLER} (see~\Cref{fig:8b_bler_distribution_alpha}), which ultimately degrades throughput.

\textbf{Setting $\boldsymbol{\alpha}$ too high (e.g., $\boldsymbol{\alpha = 2}$)} has the opposite effect, as it heavily penalizes failed transmissions. The resulting policy promotes very conservative \ac{MCS} selection, as shown in~\Cref{fig:8a_mcs_distribution_alpha} (in green). Selecting overly low \ac{MCS} indices leads to transmissions targeting a spectral efficiency lower than the effective channel capacity, thus carrying fewer information bits—but with high reliability. As a result, the policy improves transmission reliability, as evident from the lower \ac{BLER} distribution in~\Cref{fig:8b_bler_distribution_alpha}, at the cost of reduced throughput and spectral efficiency across all benchmarks, as shown in~\Cref{fig:7a_throughput,fig:7b_se}, respectively.

More precisely, we can identify three clear trends:
\begin{itemize}
    \item \textit{BLER consistently decreases as $\alpha$ increases}, dropping from double-digit percentages at $\alpha = 0$ to low single-digit percentages at $\alpha = 2$.
    \item \textit{Average MCS index selection declines with increasing $\alpha$}, resulting in lower modulation orders and code rates, and reduced data transmission.
    \item \textit{Spectral efficiency decreases monotonically with increasing $\alpha$}, particularly in high-capacity scenarios such as B4, where it drops from approximately $+40\%$ at $\alpha = 0$ to about $+15\%$ at $\alpha = 2$.
\end{itemize}
Thus, increasing $\alpha$ effectively trades off spectral efficiency and data rate for improved reliability.

\textbf{Setting $\boldsymbol{\alpha}$ to optimize throughput} requires a moderate penalty that balances spectral efficiency and reliability. In our evaluations, this balance occurs near $\alpha = 0.5$. At this value, the policy reduces retransmissions without overly suppressing \ac{MCS} aggressiveness, thereby improving throughput. This behavior is observed in most multi-cell scenarios from~\Cref{table:benchmarks}—e.g., B2(a/b), B3, B4, and B5—as well as in benchmark B1, where performance losses at $\alpha = 0$ become modest gains at $\alpha = 0.5$. Hence, the throughput curve exhibits a \textbf{concave profile}, peaking near $\alpha \approx 0.5$.

Consequently, system designers should select $\alpha$ sufficiently high to mitigate excessive retransmissions—typically around 0.5 in these scenarios—without unduly compromising capacity through overly conservative behavior.

\begin{figure*}
    \centering
    \includegraphics[width=0.8\textwidth]{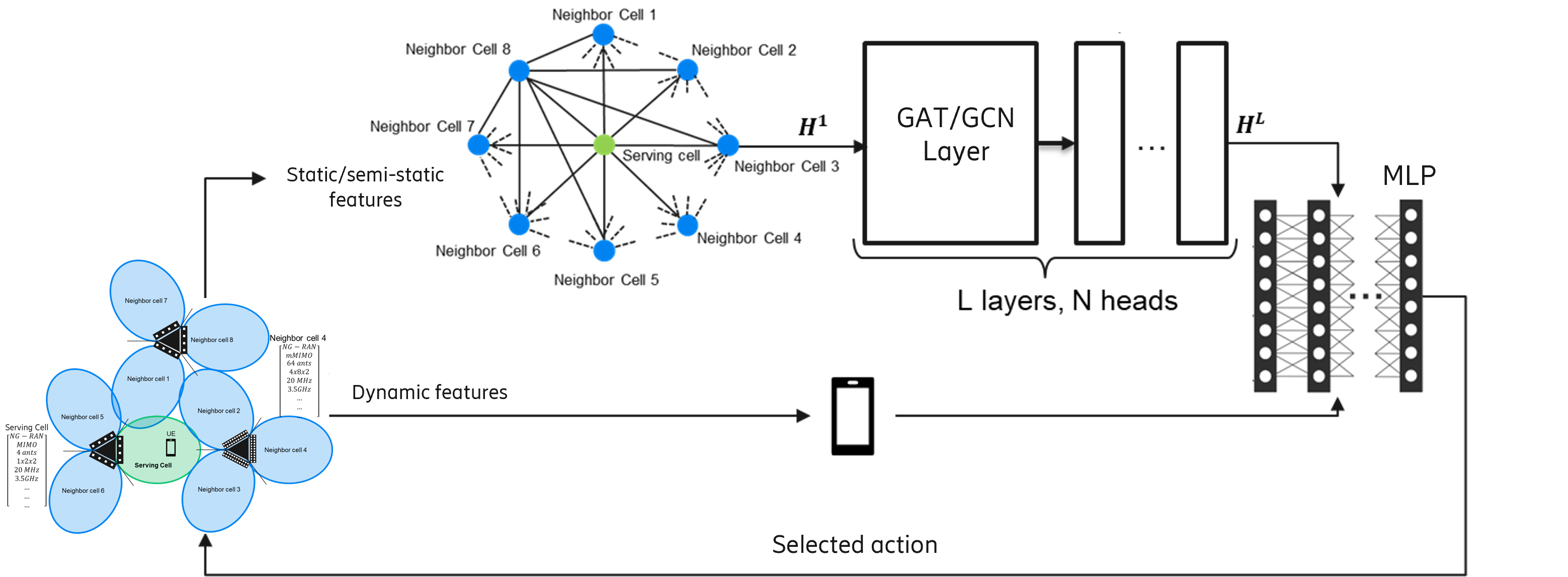}
    \caption{\textbf{Overview of the proposed architecture.} The \ac{UE}-centric network view (bottom left) is used to construct a graph where nodes represent the serving and neighboring cells. Static and semi-static features (e.g., configuration parameters) are processed by a GNN (e.g., GCN/GAT), implemented as a feedforward architecture with $L$ layers (e.g., GCN or GAT). Dynamic features (e.g., \ac{UE} measurements) are provided directly to an \ac{MLP}. Both the GNN and MLP branches are jointly trained via backpropagation, enabling the model to capture both topological cell relations and instantaneous context.}
    \label{fig:graph_enc}
\end{figure*}

\begin{figure*}
    \centering
    \includegraphics[width=0.9\textwidth]{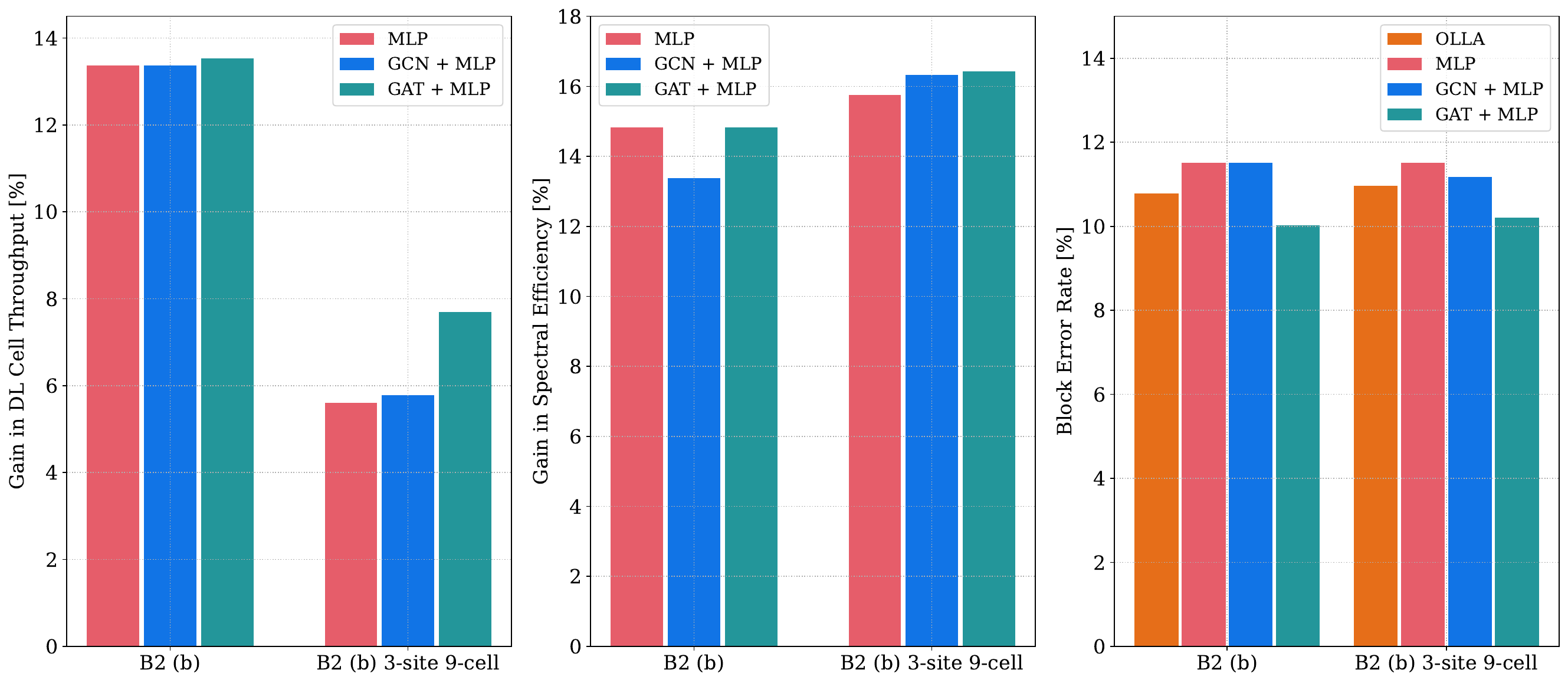}
        \caption{Comparison between MLP-only and GNN-based models (\ac{GCN} + \ac{MLP} and \ac{GAT} + \ac{MLP}, as per~\Cref{fig:graph_enc}) for two versions of benchmark \textbf{B2(b)}. Within each plot, the left-most group presents results for \emph{1-site 1-cell} version, while the right-most group refers to a denser \emph{3-site 9-cell} deployment. Images from left to right show average downlink cell throughput and spectral efficiency gain with respect to \ac{OLLA}, as well as \ac{BLER}, respectively.}
    \label{fig:graph_results}
\end{figure*}


\subsection{Impact of Graph-Based Approach} \label{subsec8:graphs}

We next analyze how using graph representations to encode static and semi-static network information—capturing both site/cell attributes and inter-site/cell relationships within a RAN topology—can improve the robustness and transferability of \ac{RL} models across diverse deployment scenarios. Our findings, presented in~\Cref{fig:graph_results}, indicate that attention-based graphs provide benefits even in relatively small network deployments (e.g., 9 cells), although the gains diminish in very small deployments (e.g., 3 cells), where there is insufficient structural diversity to exploit.

The results in~\Cref{fig:graph_results} are obtained by encoding the static and semi-static network information outlined in~\Cref{subsec7:static_information}, such as cell type, number of antennas, downlink transmit power, and average cell utilization, using a \ac{GNN}. The overall model architecture, illustrated in~\Cref{fig:graph_enc}, consists of a cascade of a \ac{GNN} followed by an \ac{MLP} for Q-value estimation. The \ac{MLP} receives as input both the encoding produced by the \ac{GNN} and the raw features representing the fast dynamics of link adaptation.

As introduced in~\Cref{sec4:graphs}, the input to the \ac{GNN} consists of attribute vectors $\mathbf{h}_i$ for nodes $i$ in a graph representing the local network deployment surrounding a \ac{UE}, where each node corresponds to either the serving cell or an interfering cell. The full set of node features is denoted by the matrix $\mathbf{H}^{(0)}$, where each row corresponds to the attribute vector of a node, and $\mathbf{H}^{(l)}$ represents the encoding produced at layer $l$ of the \ac{GNN}.

We consider two \ac{GNN}-based architectures: \textit{\ac{GCN} + \ac{MLP}} and \textit{\ac{GAT} + \ac{MLP}}. In both cases, the \ac{GNN} and subsequent \ac{MLP} are trained end-to-end using backpropagation, minimizing Q-value prediction error following the procedure described in~\Cref{sec:training_setup}. While we tested all five benchmarks from~\Cref{table:benchmarks}, \Cref{fig:graph_results} presents results only for benchmark B2(b), as similar conclusions were observed across benchmarks. For this analysis, we also include an extended version of B2 with three sites and nine cells.

We compare these graph-based models with the \ac{MLP}-only architecture introduced in~\Cref{sec8c:general_results}, which consists of a 6-layer \ac{MLP} with 256 neurons per layer and no \ac{GNN}-based encoding. All models are trained with a robustness weight of $\alpha = 0.5$ in~\Cref{eqn:reward}, and we use \acp{GNN} with a single layer due to the fully connected graph topology and to reduce computational complexity.

\Cref{fig:graph_results} presents results for all approaches in terms of: (a) throughput gain relative to the \ac{OLLA} baseline, (b) spectral efficiency gain relative to the \ac{OLLA} baseline, and (c) average experienced \ac{BLER}. We observe that, while \ac{GNN}-based and \ac{MLP}-only models exhibit similar performance in small deployments (e.g., B2(b)), the \ac{GAT} model achieves approximately 30\% higher throughput than both the \ac{GNN}-based and \ac{MLP}-only models in the extended 9-cell deployment. This suggests that incorporating attention mechanisms, such as those in the \ac{GAT} architecture, becomes increasingly beneficial in larger network surrounding, where the topology and interrelations between radio sites/cells become more complex and diverse. 

Although our analysis is limited to 9-cell scenarios to reduce simulation time and relies on a constrained set of randomization parameters, the results indicate that graph-based encodings—particularly attention-based ones—may be instrumental for real-world deployments, where cells often operate in highly diverse, irregular, and heterogeneous environments. This is the focus of our ongoing research.


\section{Conclusions and Future Extensions}\label{sec:conclusions}

This work addressed a fundamental challenge in applying deep \ac{RL} to \acp{RAN}: achieving robust generalization across diverse and dynamic wireless environments. Although \ac{RL} shows great promise for automating complex \ac{RAN} control functions—such as those involved in \ac{RRM} and network orchestration—its deployment in real-world \acp{RAN} remains limited by the difficulty of transferring models trained under specific conditions to heterogeneous environments characterized by varying deployment scenarios, user behaviors, and channel dynamics.

To address this limitation, we introduced a comprehensive framework to improve generalization in \ac{RL}-based \ac{RAN} applications. We formally defined \ac{RL} generalization in the context of \acp{RAN} and identified key challenges, including partial observability and the diversity and non-stationarity of radio environments. We then proposed three core enablers for developing generalizable \ac{RL} policies: (i) improved state abstraction tailored to RAN-specific characteristics; (ii) diversification of training environments to expose models to a wide range of deployment scenarios and radio environment conditions; and (iii) a scalable architecture for distributed learning, supporting large-scale data collection across simulated and real network nodes.

The proposed \ac{AI} architecture supports not only sim2sim generalization but also alignment with the structural and operational requirements of real-world \acp{RAN}, making it suitable for deployment in live networks. This capability enables models to be trained directly on field data collected across the network, thereby substantially mitigating performance discrepancies induced by the sim2real gap. In this setting, models initially trained in simulation can serve as a warm-up phase before fine-tuning on real-world data gathered from distributed network nodes.

We validated the feasibility and effectiveness of the proposed approach by developing a novel \ac{RL} algorithm for link adaptation. Using a high-fidelity system-level simulator compliant with \ac{5G} \ac{NR} standards, we demonstrated that \ac{RL} algorithms trained with the proposed generalization enablers outperform both fine-tuned models and state-of-the-art heuristics when evaluated across diverse and previously unseen environments. These results highlight the potential of zero-shot generalization in realistic network settings, thereby reducing the need for frequent retraining.

Overall, this work presents a concrete step toward scalable and generalizable \ac{AI} solutions for wireless networks. By enhancing policy robustness and simplifying lifecycle management, it paves the way for the practical integration of \ac{RL} solutions into next-generation AI-native RANs.

\clearpage
\bibliographystyle{assets/plainnat}
\bibliography{references}

\newpage

\section{Supplementary Information} \label{Supplementary}
\renewcommand{\thefigure}{S\arabic{figure}}
\setcounter{figure}{0}


\subsection{Hyperparameters}\label{app:hyperparams}

For completeness, the hyperparameters used in the distributed \ac{RL} training, as described in~\Cref{sec:architecture}, are summarized in~\Cref{table:hyper_param}.

\begin{table}[th!]
	\caption{\ac{RL} training hyperparameters.}
	\centering
	\begin{tabular}{l l}
		\toprule[1pt]\midrule[0.3pt]
		\textbf{Hyperparameter} & \textbf{Value} \\ [0.5ex]
		\midrule
		Discount factor ($\gamma$) & 1.0  \\
        $n$-step & 1 \\
        Actor's local buffer capacity & 500 \\
        Model update interval & Every 512 gradient updates \\
        Target network update method & Hard \\
		Target network update interval & Every 2500 gradient updates \\
        Warm-up phase & 45,000 samples \\
    	\midrule
        Loss function & MSE / Huber \\
        Optimizer & Adam~\cite{KiB:17} \\
        Learning rate & 0.00005 \\ 
        $\beta_{1}$ (Adam momentum term) & 0.9 \\
        $\beta_{2}$ (Adam second moment term) & 0.999 \\
        $\epsilon$ (Adam numerical stability) & 0.0001 \\
        Weight decay & 0.02/512 \\
        Maximum gradient norm & 20 \\
        Batch size & 512 \\
        Prefetched batches & 16 \\
    	\midrule
        Number of replay shards & 4 \\
        Replay buffer capacity & 4,000,000 \\
        Prioritization exponent ($\alpha$) & 0.6 \\
        Importance sampling exponent ($\beta$) & 0.4 \\
		\midrule[0.3pt]\bottomrule[1pt]
	\end{tabular}\label{table:hyper_param}
\end{table}


\subsection{Compute Resources} \label{app:compute_resources}

All experiments were conducted on a \ac{HPC} cluster. The primary compute node was equipped with an NVIDIA A100-PCIE-40GB GPU, featuring 40 GB of high-bandwidth HBM2 memory, and 48 CPU cores for managing actor processes, the learner, and the replay memory. The replay memory was partitioned into four independently prioritized shards, each assigned to a dedicated CPU core to allow parallel access and storage. The actors, learner, and replay memory were all co-located on this primary node to minimize intra-node communication latency. Each experiment utilized 40 actors, with each actor spawning two threads to concurrently interact with 14 simulation instances. Simulation environments were executed on multiple separate compute nodes comprising 560 CPU cores in total. Each simulation instance ran as an independent process and communicated with its respective actor via ZeroMQ, allowing scalable and distributed environment interaction. Job scheduling and resource allocation across the cluster were managed by the \ac{LSF}, a workload management system designed for distributed \ac{HPC} environments. \ac{LSF} handled job submission, queueing, monitoring, and allocation of compute nodes based on the specified resource constraints.

\begin{figure}
    \centering
    \includegraphics[width=0.9\columnwidth]{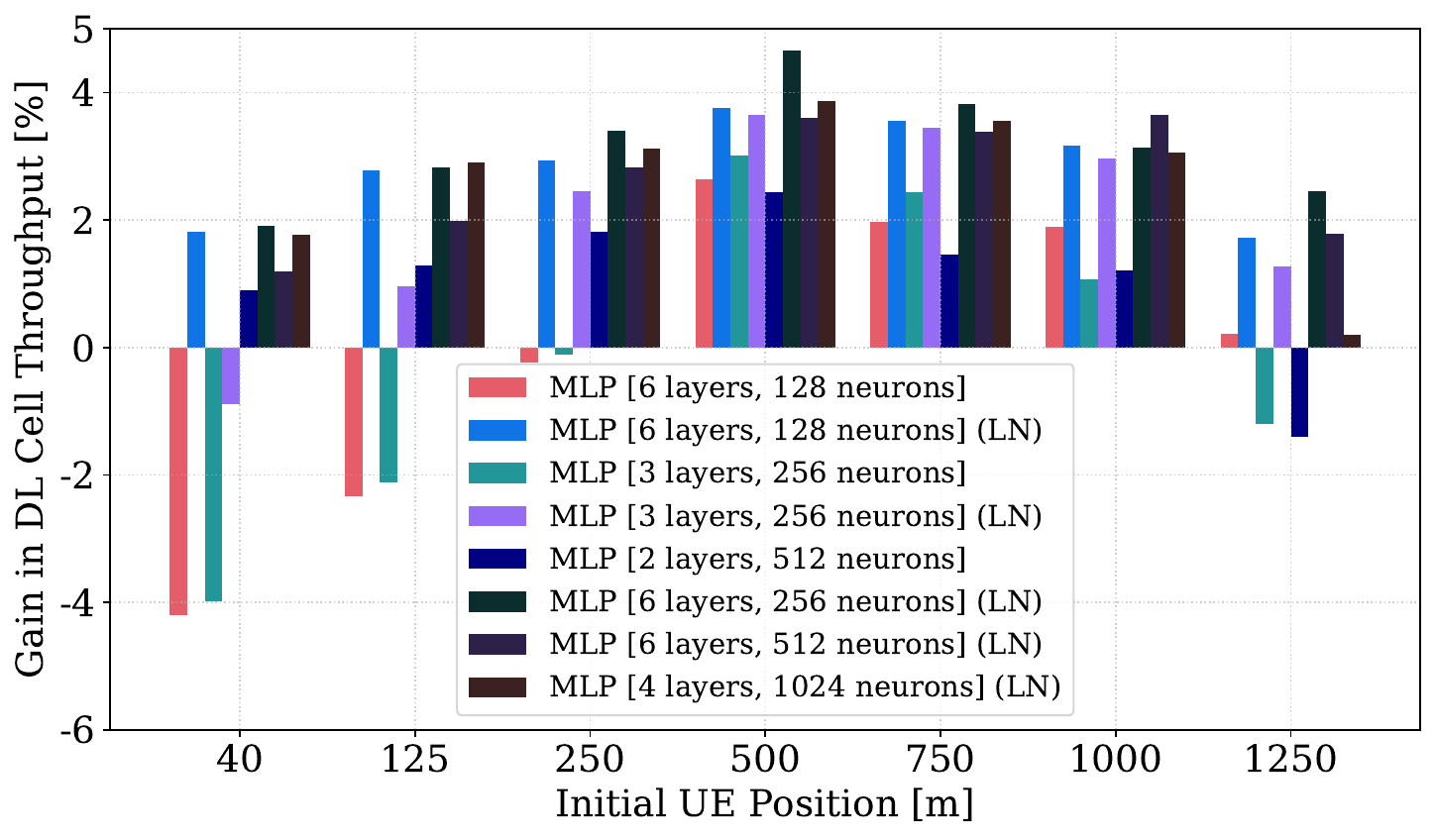}
    \caption{\textbf{Impact of neural network architecture on downlink cell throughput gains for varying \ac{UE} positions in benchmark B1.} Relative gain in mean downlink cell throughput (in percent) as a function of a \ac{UE}’s initial position (30--1{,}250\,m) for eight \ac{MLP} variants. Network configurations vary in depth (2, 3, 4 or 6 hidden layers), width (128, 256, 512 or 1{,}024 neurons per layer) and the presence of layer normalization (LN).}
    \label{fig:SCSU_model_comp}
\end{figure}


\subsection{Neural Network Architectures} \label{app:network_architecture} 

\Cref{fig:SCSU_model_comp} further analyzes performance variations across different neural network architectures for benchmark B1, where \ac{UE} distances to the access node range from approximately \qty{40}{\meter} to the cell edge (about \qty{1250}{\meter}). All architectures employing \ac{LN}~\cite{BKH:16} consistently achieve positive throughput gains throughout the entire cell radius, highlighting the essential role of \ac{LN} in maintaining stability. Among these, the 6-layer, 256-unit \ac{LN} model demonstrates the most stable and reliable performance: it provides a gain of approximately 2\% near the cell center, peaks at around 4–5\% in the heavily loaded mid-cell region (\qtyrange{250}{750}{\metre}), and maintains about 2\% gain at the cell edge. Wider \ac{LN} models, including the 6-layer, 512-unit and 4-layer, 1024-unit networks, occasionally surpass this mid-cell performance but show slight reductions near the cell edge. In contrast, models lacking \ac{LN} or featuring smaller capacities—such as the 3-layer, 256-unit model—exhibit negative throughput gains at various locations, indicating underfitting or instability when \ac{SINR} deviates from typical training conditions.

\Cref{tab:model_benchmarks} confirms that \ac{LN} generally improves spectral efficiency and often increases throughput, contributing an additional 1–2 percentage points in spectral efficiency and up to 1 percentage point in throughput compared to equivalent models without \ac{LN}. Gains are most notable at moderate complexity, particularly with the 6-layer, 256-unit \ac{LN} architecture (approximately 340k parameters), which achieves leading results across critical metrics: B2-MIMO throughput (+8.50\%), spectral efficiency (+8.07\%), B3 throughput (+8.39\%), and spectral efficiency (+7.46\%). This model consistently performs within approximately 2 percentage points of the best-performing models in other scenarios.

Larger architectures offer specific advantages, such as the 6-layer, 512-unit \ac{LN} model, which achieves the highest spectral efficiency in scenario B4 (+30.74\%) and B2-mMIMO (+13.92\%), and the 2-layer, 512-unit network, which delivers the best throughput in scenario B4 (+21.34\%). However, these larger models typically underperform in other scenarios and require significantly longer training times (4–12 additional hours). Conversely, smaller models (fewer than 200k parameters), such as the 6-layer, 128-unit model without \ac{LN}, reduce computational load but fail to consistently deliver positive gains.

In summary, the 6-layer, 256-unit \ac{LN} architecture offers consistently positive throughput gains across all user positions (see~\Cref{fig:SCSU_model_comp}) and achieves the most balanced performance across deployment scenarios, with reasonable computational cost (see~\Cref{tab:model_benchmarks}). While wider models may be justified in scenarios demanding peak spectral efficiency—such as high-mobility macro-cell environments (e.g., scenario B4)—for general deployment, the 6-layer, 256-unit \ac{LN} architecture represents the most efficient and practical choice for \ac{5G} link adaptation applications.

\begin{table*}[th!]
    \caption{Performance comparison of \ac{MLP} architectures relative to the LA-OLLA baseline (10\% BLER target, typical in 5G systems). Gains in average cell throughput (Throughput) and spectral efficiency (Spec. Eff.) are shown.}
    \centering
    \resizebox{\textwidth}{!}{
    \begin{tabular}{@{}l r r cc cc cc cc cc@{}}
        \toprule
        & & & \multicolumn{2}{c}{\textbf{B2 MIMO}} & \multicolumn{2}{c}{\textbf{B2 mMIMO}} & \multicolumn{2}{c}{\textbf{B3}} & \multicolumn{2}{c}{\textbf{B4}} & \multicolumn{2}{c}{\textbf{B5}} \\
        \cmidrule(lr){4-5} \cmidrule(lr){6-7} \cmidrule(lr){8-9} \cmidrule(lr){10-11} \cmidrule(lr){12-13}
        \textbf{Architecture} & \textbf{\#Params} & \textbf{Train. Time} & Throughput & Spec. Eff. & Throughput & Spec. Eff. & Throughput & Spec. Eff. & Throughput & Spec. Eff. & Throughput & Spec. Eff. \\
        \midrule
        6L$\times$128             & 88k     & 26h 21m  & 8.06\%   & 6.71\%  & 13.24\% & 13.12\%  & 4.52\% & 5.38\%  & 21.35\% & 27.65\%  & 3.10\% & 3.77\% \\
        6L$\times$128 (LN)        & 88k     & 33h 46m  & 8.05\%   & 7.96\%  & 12.90\% & 13.32\%  & 7.23\% & 6.89\%  & 18.69\% & 30.22\%  & 5.96\% & 5.30\% \\
        3L$\times$256             & 144k    & 31h 50m  & 7.32\%   & 6.65\%  & 12.75\% & 13.17\%  & 6.96\% & 7.35\%  & 20.57\% & 28.47\%  & \textbf{6.60\%} & \textbf{5.83\%} \\
        3L$\times$256 (LN)        & 144k    & 30h 16m  & 8.40\%   & 8.02\%  & \textbf{13.28\%} & 13.79\%  & 6.91\% & 6.66\%  & 20.13\% & 29.67\%  & 5.83\% & 5.42\% \\
        2L$\times$512             & 285k    & 28h 59m  & 7.45\%   & 6.71\%  & 12.64\% & 13.14\%  & 6.99\% & 6.88\%  & \textbf{21.34\%} & 28.82\%  & 5.96\% & 5.49\% \\
        6L$\times$256 (LN)        & 340k    & 34h 14m  & \textbf{8.50\%} & \textbf{8.07\%} & 12.98\% & 13.83\%  & \textbf{8.39\%} & \textbf{7.46\%} & 19.32\% & 29.59\%  & 6.44\% & 5.70\% \\
        6L$\times$512 (LN)        & 1.34M   & 38h 41m  & 8.37\%   & 7.99\%  & 13.24\% & \textbf{13.92\%}  & 6.65\% & 6.67\%  & 19.19\% & \textbf{30.74\%} & 4.78\% & 5.02\% \\
        4L$\times$1024 (LN)        & 3.20M   & 50h  08m  & 8.47\%   & 7.66\%  & 13.16\% & 13.63\%  & 6.60\% & 6.22\%  & 20.18\% & 29.39\% & 5.03\% & 5.01\% \\
        \bottomrule
    \end{tabular}
    }
    \label{tab:model_benchmarks}
\end{table*}

\end{document}